\documentclass[10pt,twocolumn,letterpaper]{article}

\usepackage{cvpr}
\usepackage{times}
\usepackage{epsfig}
\usepackage{graphicx}
\usepackage{amsmath}
\usepackage{amssymb}
\usepackage{multirow}
\usepackage{subfigure}

\usepackage[breaklinks=true,bookmarks=false]{hyperref}

\cvprfinalcopy 

\def\cvprPaperID{****} 

\ifcvprfinal\fi
\begin{document}

\title{ECA-Net: Efficient Channel Attention for Deep Convolutional Neural Networks}

\author{Qilong Wang$^1$, Banggu Wu$^1$, Pengfei Zhu$^1$, Peihua Li$^2$, Wangmeng Zuo$^3$, Qinghua Hu$^{1,}$\thanks{Qinghua Hu is the corresponding author. 
		\newline Email: \{qlwang, wubanggu, huqinghua\}@tju.edu.cn.	 The work was supported by the National Natural Science Foundation of China (Grant No. 61806140, 61876127, 61925602, 61971086, U19A2073, 61732011), Major Scientific Research Project of Zhejiang Lab (2019DB0ZX01). Q. Wang was supported by National Postdoctoral Program for Innovative Talents.} \\ 
	$^1$ Tianjin Key Lab of Machine Learning, College of Intelligence and Computing, Tianjin University, China\\
	$^2$ Dalian University of Technology, China \,\,\,\,\, $^3$ Harbin Institute of Technology, China\\	
}

\maketitle
\thispagestyle{empty}

\begin{abstract}
	Recently, channel attention mechanism has demonstrated to offer great potential in improving the performance of deep convolutional neural networks (CNNs). However, most existing methods dedicate to developing more sophisticated attention modules for achieving better performance, which inevitably increase model complexity. To overcome the paradox of performance and complexity trade-off, this paper proposes an Efficient Channel Attention (ECA) module, which only involves a handful of parameters while bringing clear performance gain. By dissecting the channel attention module in SENet, we empirically show avoiding dimensionality reduction is important for learning channel attention, and appropriate cross-channel interaction can preserve performance while significantly decreasing model complexity. Therefore, we propose a local cross-channel interaction strategy without dimensionality reduction, which can be efficiently implemented via $1D$ convolution. Furthermore, we develop a method to adaptively select kernel size of $1D$ convolution, determining coverage of local cross-channel interaction. The proposed ECA module is efficient yet effective, e.g., the parameters and computations of our modules against backbone of ResNet50 are 80 vs. 24.37M and 4.7e-4 GFLOPs vs. 3.86 GFLOPs, respectively, and the performance boost is more than 2\% in terms of Top-1 accuracy. We extensively evaluate our ECA module on image classification, object detection and instance segmentation with backbones of ResNets and MobileNetV2. The experimental results show our module is more efficient while performing favorably against its counterparts.
\end{abstract}

\section{Introduction}

Deep convolutional neural networks (CNNs) have been widely used in computer vision community, and have achieved great progress in a broad range of tasks, \eg, image classification, object detection and semantic segmentation. Starting from the groundbreaking AlexNet \cite{nips2012cnn}, many researches are continuously investigated to further improve the performance of deep CNNs~\cite{Simonyan15,Szegedy_2015_CVPR,He_2016_CVPR,Huang_2017_CVPR,LiXWZ17,LiYanghao_2017_ICCV,Wang_2018_CVPR}. Recently, incorporation of channel attention into convolution blocks has attracted a lot of interests, showing great potential in performance improvement~\cite{SENet18,Woo_2018_ECCV,DBLP:conf/nips/HuSASV18,A2NIPS18,Gao_2019_CVPR,DBLP-1901-01493,Fu_2019_CVPR}. One of the representative methods is squeeze-and-excitation networks (SENet)~\cite{SENet18}, which learns channel attention for each convolution block, bringing clear performance gain for various deep CNN architectures.

\begin{figure}[t]
	\centering
	\includegraphics[width=0.82\columnwidth]{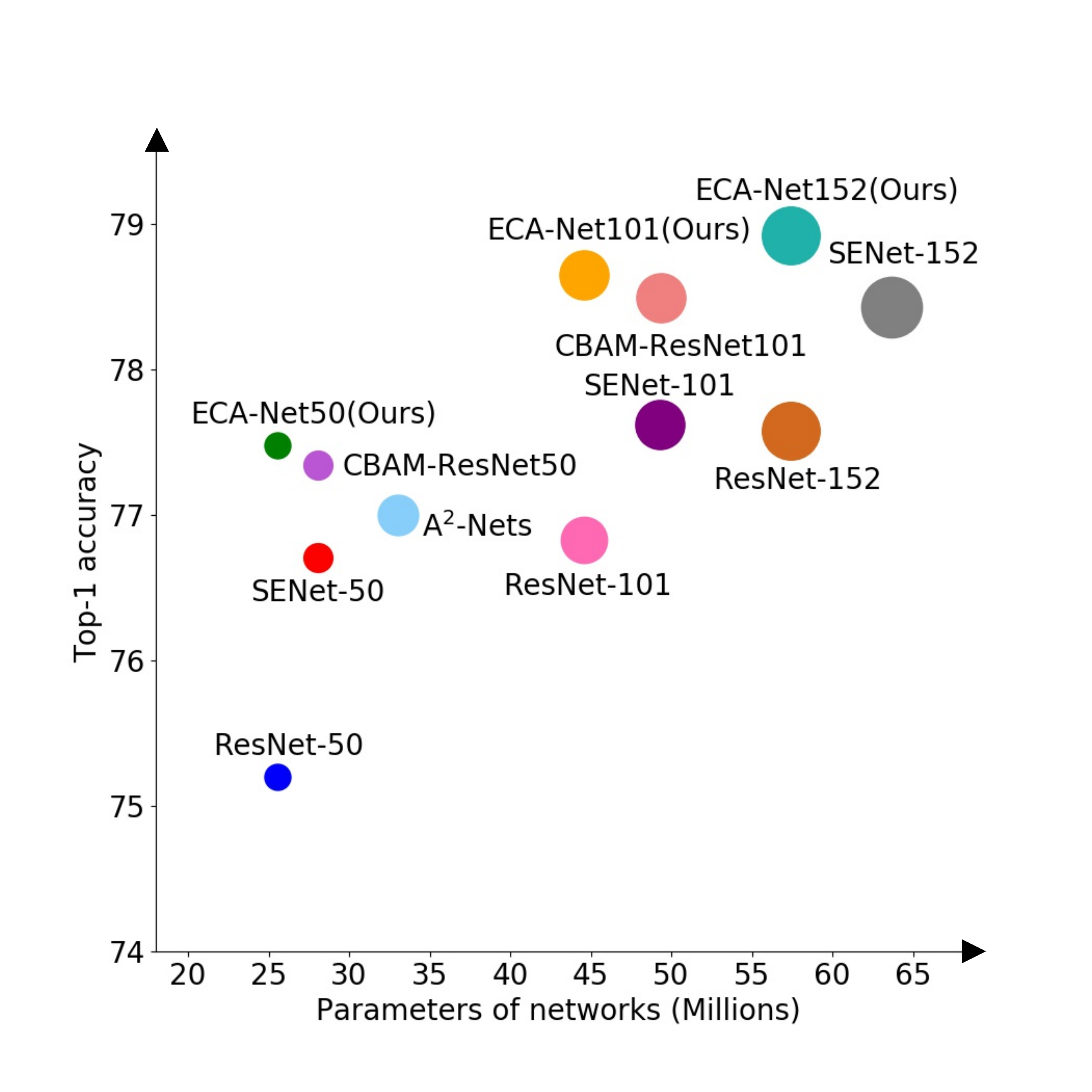}
	\caption{Comparison of various attention modules (i.e., SENet~\cite{SENet18}, CBAM~\cite{Woo_2018_ECCV},  $A^{2}$-Nets~\cite{A2NIPS18} and ECA-Net) using ResNets~\cite{He_2016_CVPR} as backbone models in terms of classification accuracy, network parameters and FLOPs, indicated by radiuses of circles. Note that our ECA-Net obtains higher accuracy while having less model complexity.}
	\label{fig:motivation}
\end{figure}

\begin{figure}[t]
	\centering
	\includegraphics[width=0.9\linewidth]{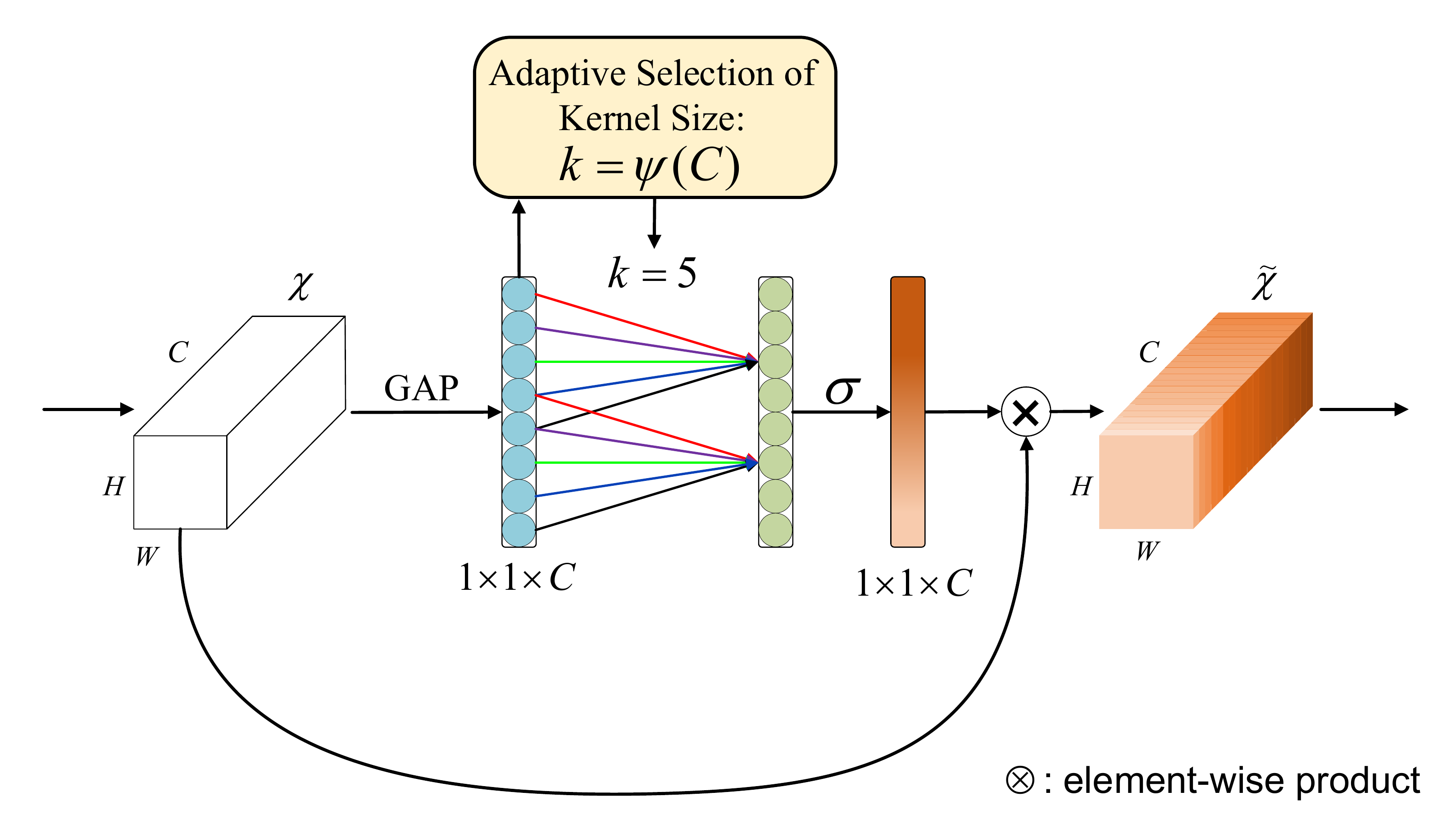}
	\caption{Diagram of our efficient channel attention (ECA) module. Given the aggregated features obtained by global average pooling (GAP), ECA generates channel weights by performing a fast $1D$ convolution of size $k$, where $k$ is adaptively determined via a mapping of channel dimension $C$.}
	\label{fig:method}
\end{figure}

Following the setting of squeeze (i.e., feature aggregation) and excitation (i.e., feature recalibration) in SENet~\cite{SENet18}, some researches improve SE block by capturing more sophisticated channel-wise dependencies~\cite{Woo_2018_ECCV,A2NIPS18,Gao_2019_CVPR,Fu_2019_CVPR} or by combining with additional spatial attention~\cite{Woo_2018_ECCV,DBLP:conf/nips/HuSASV18,Fu_2019_CVPR}. Although these methods have achieved higher accuracy, they often bring higher model complexity and suffer from heavier computational burden. Different from the aforementioned methods that achieve better performance at the cost of higher model complexity, this paper focuses instead on a question: \emph{Can one learn effective channel attention in a more efficient way?}

To answer this question, we first revisit the channel attention module in SENet. Specifically, given the input features, SE block first employs a global average pooling for each channel independently, then two fully-connected (FC) layers with non-linearity followed by a Sigmoid function are used to generate channel weights. The two FC layers are designed to capture non-linear cross-channel interaction, which involve dimensionality reduction for controlling model complexity. Although this strategy is widely used in subsequent channel attention modules~\cite{Woo_2018_ECCV,DBLP:conf/nips/HuSASV18,Gao_2019_CVPR}, our empirical studies show dimensionality reduction brings side effect on channel attention prediction, and it is inefficient and unnecessary to capture dependencies across all channels.   

Therefore, this paper proposes an \emph{Efficient Channel Attention} (ECA) module for deep CNNs, which avoids dimensionality reduction and captures cross-channel interaction in an efficient way. As illustrated in Figure~\ref{fig:method}, after channel-wise global average pooling without dimensionality reduction, our ECA captures local cross-channel interaction by considering every channel and its $k$ neighbors. Such method is proven to guarantee both efficiency and effectiveness. Note that our ECA can be efficiently implemented by fast $1D$ convolution of size $k$, where kernel size $k$ represents the coverage of local cross-channel interaction, i.e., how many neighbors participate in attention prediction of one channel. To avoid manual tuning of $k$ via cross-validation, we develop a method to adaptively determine $k$, where  coverage of interaction (i.e., kernel size $k$) is proportional to channel dimension. As shown in Figure~\ref{fig:motivation} and Table~\ref{table2}, as opposed to the backbone models~\cite{He_2016_CVPR}, deep CNNs with our ECA module (called ECA-Net) introduce very few additional parameters and negligible computations, while bringing notable performance gain. For example, for ResNet-50 with 24.37M parameters and 3.86 GFLOPs, the additional parameters and computations of ECA-Net50  are 80 and 4.7e-4 GFLOPs, respectively; meanwhile, ECA-Net50 outperforms ResNet-50 by 2.28\%  in terms of Top-1 accuracy. 

Table~\ref{Atribute} summarizes existing attention modules in terms of whether channel dimensionality reduction (DR), cross-channel interaction and lightweight model, where we can see that our ECA module learn effective channel attention by avoiding channel dimensionality reduction while capturing cross-channel interaction in an extremely lightweight way. To evaluate our method, we conduct experiments on ImageNet-1K~\cite{imagenet_cvpr09} and MS COCO~\cite{lin2014microsoft} in a variety of tasks using different deep CNN architectures. 

The contributions of this paper are summarized as follows. (1) We dissect the SE block and empirically demonstrate avoiding dimensionality reduction and appropriate cross-channel interaction are important to learn effective and efficient channel attention, respectively. (2) Based on above analysis, we make an attempt to develop an extremely lightweight channel attention module for deep CNNs by proposing an \emph{Efficient Channel Attention} (ECA), which increases little model complexity while bringing clear improvement. (3) The experimental results on ImageNet-1K and MS COCO demonstrate our method has lower model complexity than state-of-the-arts while achieving very competitive performance.

\begin{table}[t]
	\centering
	\footnotesize
	\smallskip
	\begin{tabular}{l|c|c|c}
		\hline
		Model                              & No DR     & Cross-channel Interaction & Lightweight \\
		\hline
		SENet~\cite{SENet18}               & $\times$  & $\surd$           & $\textendash$    \\
		CBAM~\cite{Woo_2018_ECCV}          & $\times$  & $\surd$           & $\times$    \\
		GE-$\theta^-$ ~\cite{DBLP:conf/nips/HuSASV18}                     & $\surd$   & $\times$          & $\surd$     \\
		GE-$\theta$~\cite{DBLP:conf/nips/HuSASV18}                        & $\surd$   & $\times$          & $\times$    \\
		GE-$\theta^+$~\cite{DBLP:conf/nips/HuSASV18}                        & $\times$   & $\surd$          & $\times$    \\
		$A^{2}$-Net~\cite{A2NIPS18}        & $\times$  & $\surd$           & $\times$    \\
		GSoP-Net~\cite{Gao_2019_CVPR}      & $\times$  & $\surd$           & $\times$    \\
		\hline
		ECA-Net (Ours)                            & $\surd$   & $\surd$           & $\surd$    \\
		\hline
	\end{tabular}
	\smallskip
	\caption{Comparison of existing attention modules in terms of whether no channel dimensionality reduction (No DR), cross-channel interaction and less parameters than SE (indicated by lightweight) or not.}
	\label{Atribute}
\end{table}

\section{Related Work}
Attention mechanism has proven to be a potential means to enhance deep CNNs. SE-Net~\cite{SENet18} presents for the first time an effective mechanism to learn channel attention and achieves promising performance. Subsequently, development of attention modules can be roughly divided into two directions: (1) enhancement of feature aggregation; (2) combination of channel and spatial attentions. Specifically, CBAM~\cite{Woo_2018_ECCV} employs both average and max pooling to aggregate features. GSoP~\cite{Gao_2019_CVPR} introduces a second-order pooling for more effective feature aggregation. GE~\cite{DBLP:conf/nips/HuSASV18} explores spatial extension using a depth-wise convolution~\cite{DBLP:conf/cvpr/Chollet17} to aggregate features.  CBAM~\cite{Woo_2018_ECCV} and scSE~\cite{DBLP:journals/tmi/RoyNW19} compute spatial attention using a $2D$ convolution of kernel size $k \times k$, then combine it with channel attention. Sharing similar philosophy with Non-Local (NL) neural networks~\cite{Wang_2018_CVPR},
GCNet~\cite{Cao_2019_ICCV_Workshops} develops a simplified NL network and integrates with the SE block, resulting in a lightweight module to model long-range dependency. Double Attention Networks ($A^{2}$-Nets)~\cite{A2NIPS18} introduces a novel relation function for NL blocks for image or video recognition. Dual Attention Network (DAN)~\cite{Fu_2019_CVPR} simultaneously considers NL-based channel and spatial attentions for semantic segmentation. However, most above NL-based attention modules can only be used in a single or a few convolution blocks due to their high model complexity. Obviously, all of the above methods focus on developing sophisticated attention modules for better performance. Different from them, our ECA aims at learning effective channel attention with low model complexity.

Our work is also related to efficient convolutions, which are designed for lightweight CNNs. Two widely used efficient convolutions are group convolutions~\cite{Zhang_2017_ICCV,XieGDTH17,Ioannou_2017_CVPR} and depth-wise separable convolutions~\cite{DBLP:conf/cvpr/Chollet17,DBLP:conf/cvpr/SandlerHZZC18,DBLP:conf/cvpr/ZhangZLS18,DBLP:conf/eccv/MaZZS18}. As given in Table~\ref{table:channel_keep}, although these efficient convolutions involve less parameters, they show little effectiveness in attention module. Our ECA module aims at capturing local cross-channel interaction, which shares some similarities with channel local convolutions~\cite{DBLP:conf/cvpr/Zhang18} and channel-wise convolutions~\cite{DBLP:conf/nips/GaoWJ18}; different from them, our method investigates a $1D$ convolution with adaptive kernel size to replace FC layers in channel attention module. Comparing with group and depth-wise separable convolutions, our method achieves better performance with lower model complexity.

\section{Proposed Method}
In this section, we first revisit the channel attention module in SENet~\cite{SENet18} (i.e., SE block). Then, we make a empirical diagnosis of SE block by analyzing effects of dimensionality reduction and cross-channel interaction. This motivates us to propose our ECA module. In addition, we develop a method to adaptively determine parameter of our ECA, and finally show how to adopt it for deep CNNs.

\begin{table}[t]
	\centering
	\footnotesize
	\begin{tabular}{lllll}
		\hline
		Methods & Attention & \#.Param. & Top-1 & Top-5 \\
		\hline
		Vanilla  & N/A & 0 & 75.20 & 92.25 \\
		SE  & $ \sigma(f_{\{\mathbf{W_{1}},\mathbf{W_{2}}\}}(\mathbf{y}))$ & $2\times C^{2}/r$ & 76.71 & 93.38 \\
		\hline
		SE-Var1  & $ \sigma(\mathbf{y})$ & $0$ & 76.00 & 92.90 \\
		SE-Var2  & $ \sigma(\mathbf{w}\odot\mathbf{y})$ & $C$ & 77.07 & 93.31 \\
		SE-Var3  & $ \sigma(\mathbf{W}\mathbf{y})$ & $C^{2}$ & 77.42 & 93.64 \\ 
		\hline
		SE-GC1  & $ \sigma(\text{GC}_{16}(\mathbf{y}))$ & $C^{2}/16$ & 76.95 & 93.47 \\
		SE-GC2  & $ \sigma(\text{GC}_{C/16}(\mathbf{y}))$ & $16\times C$ & 76.98 & 93.31 \\
		SE-GC3  & $ \sigma(\text{GC}_{C/8}(\mathbf{y}))$ & $8\times C$ & 76.96 & 93.38 \\
		\hline
		ECA-NS  & $ \sigma(\boldsymbol\omega)$ with Eq.~(\ref{fun_eca1}) & $k\times C$ & 77.35 & 93.61 \\
		ECA (Ours)  & $ \sigma(\text{C1D}_{k}(\mathbf{y}))$ & $k=3$ & \textbf{77.43} & \textbf{93.65} \\
		\hline
	\end{tabular}
	\smallskip
	\caption{Comparison of various channel attention modules using ResNet-50 as backbone model on ImageNet. \#.Param. indicates number of parameters of the channel attention module; $\odot$ indicates element-wise product; GC and C1D indicate group convolutions and $1D$ convolution, respectively; $k$ is kernel size of C1D.}\smallskip
	\label{table:channel_keep}	
\end{table}

\subsection{Revisiting Channel Attention in SE Block}
Let the output of one convolution block be $\mathcal{X} \in \mathbb{R}^{W \times H \times C}$, where $W$, $H$ and $C$ are width, height and channel dimension (i.e., number of filters). Accordingly, the weights of channels in SE block can be computed as
\begin{align}\label{classical_CA}
\boldsymbol\omega = \sigma(f_{\{\mathbf{W_{1}},\mathbf{W_{2}}\}}(g(\mathcal{X}))),
\end{align}
where $g(\mathcal{X})=\frac{1}{WH}\sum_{i=1,j=1}^{W,H}\mathcal{X}_{ij}$ is channel-wise global average pooling (GAP) and $\sigma$ is a Sigmoid function. Let $\mathbf{y}=g(\mathcal{X})$,  $f_{\{\mathbf{W_{1}},\mathbf{W_{2}}\}}$ takes the form
\begin{align}\label{fun_pw}
f_{\{\mathbf{W_{1}},\mathbf{W_{2}}\}}(\mathbf{y}) =\mathbf{W_{2}}\text{ReLU}(\mathbf{W_{1}}\mathbf{y}),
\end{align}
where ReLU indicates the Rectified Linear Unit~\cite{DBLP:conf/icml/NairH10}. To avoid high model complexity, sizes of $\mathbf{W_{1}}$ and $\mathbf{W_{2}}$ are set to $C \times (\frac{C}{r})$ and $(\frac{C}{r}) \times C$, respectively. We can see that $f_{\{\mathbf{W_{1}},\mathbf{W_{2}}\}}$ involves all parameters of channel attention block. While dimensionality reduction in Eq.~(\ref{fun_pw}) can reduce model complexity, it  destroys the direct correspondence between channel and its weight. For example, one single FC layer predicts weight of each channel using a linear combination of all channels. But Eq.~(\ref{fun_pw}) first projects channel features into a low-dimensional space and then maps them back, making correspondence between channel and its weight be indirect.

\subsection{Efficient Channel Attention (ECA) Module}
After revisiting SE block, we conduct empirical comparisons for analyzing effects of channel dimensionality reduction and cross-channel interaction on channel attention learning. According to these analyses, we propose our efficient channel attention (ECA) module.

\subsubsection{Avoiding Dimensionality Reduction} As discussed above, dimensionality reduction in Eq.~(\ref{fun_pw}) makes correspondence between channel and its weight be indirect. To verify its effect, we compare the original SE block with its three variants (i.e., SE-Var1, SE-Var2 and SE-Var3), all of which do not perform dimensionality reduction. As presented in Table~\ref{table:channel_keep}, SE-Var1 with no parameter is still superior to the original network, indicating channel attention has ability to improve performance of deep CNNs. Meanwhile, SE-Var2 learns the weight of each channel independently, which is slightly superior to SE block while involving less parameters. It may suggest that channel and its weight needs a direct correspondence while avoiding dimensionality reduction is more important than consideration of nonlinear channel dependencies. Additionally, SE-Var3 employing one single FC layer performs better than two FC layers with dimensionality reduction in SE block. All of above results clearly demonstrate avoiding dimensionality reduction is helpful to learn effective channel attention. Therefore, we develop our ECA module without channel dimensionality reduction.

\subsubsection{Local Cross-Channel Interaction}

Given the aggregated feature $\mathbf{y} \in \mathbb{R}^{C}$ without dimensionality reduction,  channel attention can be learned by
\begin{align}\label{fun_cw}
\boldsymbol\omega  = \sigma(\mathbf{W}\mathbf{y}),
\end{align}
where $\mathbf{W}$ is a $C \times C$ parameter matrix. In particular, for SE-Var2 and SE-Var3 we have 
\begin{align}\label{fun_ww}
\mathbf{W} = \left\{
\begin{array}{lr}
\mathbf{W}_{var2}=\begin{bmatrix}
w^{1,1} & \cdots & 0\\
\vdots & \ddots & \vdots  \\
0 & \dots & w^{C,C}
\end{bmatrix}, &  \\
\mathbf{W}_{var3}=\begin{bmatrix}
w^{1,1} & \cdots & w^{1,C}\\
\vdots & \ddots & \vdots  \\
w^{1,C} & \dots & w^{C,C}
\end{bmatrix}, &  
\end{array}
\right.
\end{align}
where $\mathbf{W}_{var2}$ for SE-Var2 is a diagonal matrix, involving $C$ parameters;  $\mathbf{W}_{var3}$ for SE-Var3 is a full matrix, involving $C \times C$ parameters. As shown in Eq.~(\ref{fun_ww}), the key difference is that SE-Var3 considers cross-channel interaction while SE-Var2 does not, and consequently SE-Var3 achieves better performance. This result indicates that cross-channel interaction is beneficial to learn channel attention. However, SE-Var3 requires a mass of parameters, leading to high model complexity, especially for large channel numbers. 

A possible compromise between SE-Var2 and SE-Var3 is extension of  $\mathbf{W}_{var2}$ to a block diagonal matrix, i.e.,     
\begin{align}\label{fun_block}
\mathbf{W}_{G} = \begin{bmatrix}
\mathbf{W}_{G}^{1} & \cdots & \mathbf{0}\\
\vdots & \ddots & \vdots  \\
\mathbf{0} & \dots & \mathbf{W}_{G}^{G}
\end{bmatrix},
\end{align}
where Eq.~(\ref{fun_block}) divides channel into $G$ groups each of which includes $C/G$ channels, and learns channel attention in each group independently, which captures cross-channel interaction in a local manner. Accordingly, it involves $C^{2}/G$ parameters. From perspective of convolution, SE-Var2, SE-Var3 and Eq.~(\ref{fun_block}) can be regarded as a depth-wise separable convolution, a FC layer and group convolutions, respectively. Here, SE block with group convolutions (SE-GC) is indicated by $
\sigma(\text{GC}_{G}(\mathbf{y})) = \sigma(\mathbf{W}_{G}\mathbf{y})$. However, as shown in ~\cite{DBLP:conf/eccv/MaZZS18}, excessive group convolutions will increase memory access cost and so decrease computational efficiency. Furthermore, as shown in Table~\ref{table:channel_keep}, SE-GC with varying groups bring no gain over SE-Var2, indicating it is not an effective scheme to capture local cross-channel interaction. The reason may be that SE-GC completely discards dependences among different groups. 

In this paper, we explore another method to capture local cross-channel interaction, aiming at guaranteeing both efficiency and effectiveness. Specifically, we employ a band matrix $\mathbf{W}_{k}$ to learn channel attention, and $\mathbf{W}_{k}$ has 

\begin{footnotesize}
	\begin{align}\label{fun_eca0}
	\begin{bmatrix}
	w^{1,1}  &\cdots & w^{1,k} & 0 & 0 & \cdots  & \cdots & 0 \\
	0  &w^{2,2} & \cdots & w^{2,k+1} & 0 & \cdots  & \cdots & 0 \\
	\vdots  & \vdots & \vdots  & \vdots  & \ddots & \vdots & \vdots  & \vdots\\
	0  &\cdots & 0 & 0 &\cdots  & w^{C,C-k+1} & \cdots & w^{C,C}\\
	\end{bmatrix}.
	\end{align}
\end{footnotesize}
Clearly, $\mathbf{W}_{k}$ in Eq.~(\ref{fun_eca0})  involves $k\times C$ parameters, which is usually less than those of Eq.~(\ref{fun_block}). Furthermore, Eq.~(\ref{fun_eca0}) avoids complete independence among different groups in Eq.~(\ref{fun_block}). As compared in Table~\ref{table:channel_keep}, the method in Eq.~(\ref{fun_eca0}) (namely ECA-NS) outperforms SE-GC of Eq.~(\ref{fun_block}). As for Eq.~(\ref{fun_eca0}), the weight of $y_i$ is calculated by only considering interaction between $y_i$ and its $k$ neighbors, i.e.,
\begin{align}\label{fun_eca1}
\omega_{i} = \sigma\bigg(\sum_{j=1}^{k}w_{i}^{j}y_{i}^{j}\bigg),\,\,y_{i}^{j} \in \Omega_{i}^{k},
\end{align}
where $\Omega_{i}^{k}$ indicates the set of $k$ adjacent channels of $y_{i}$. 

A more efficient way is to make all channels share the same learning parameters, i.e.,
\begin{align}\label{fun_eca2}
\omega_{i} = \sigma\bigg(\sum_{j=1}^{k}w^{j}y_{i}^{j}\bigg),\,\,y_{i}^{j} \in \Omega_{i}^{k}.
\end{align}
Note that such strategy can be readily implemented by a fast $1D$ convolution with kernel size of $k$, i.e.,
\begin{align}\label{fun_eca3}
\boldsymbol\omega = \sigma(\text{C1D}_{k}(\mathbf{y})),
\end{align}
where C1D indicates $1D$ convolution. Here, the method in Eq.~(\ref{fun_eca3}) is called by efficient channel attention (ECA) module, which only involves $k$ parameters. As presented in Table~\ref{table:channel_keep}, our ECA module with $k=3$ achieves similar results with SE-var3 while having much lower model complexity, which guarantees both efficiency and effectiveness by appropriately capturing local cross-channel interaction.

\begin{figure}[t]
	\centering
	\includegraphics[width=0.94\columnwidth]{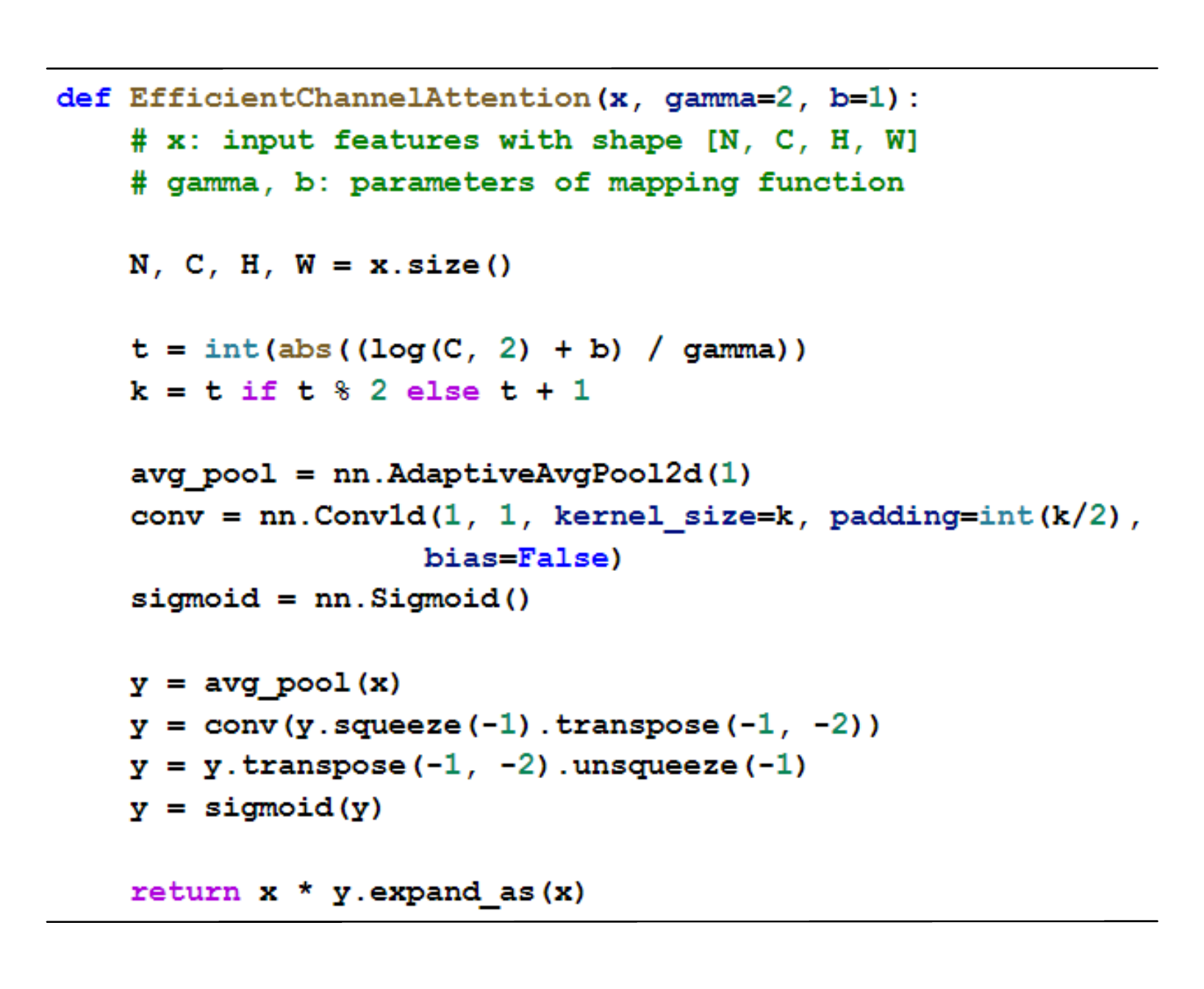}
	\caption{PyTorch code of our ECA module.}
	\label{fig:code}
\end{figure}

\subsubsection{Coverage of Local Cross-Channel Interaction}
Since our ECA module (\ref{fun_eca3}) aims at appropriately capturing local cross-channel interaction, so the coverage of interaction (i.e., kernel size $k$ of $1D$ convolution) needs to be determined. The optimized coverage of interaction  could be tuned manually for convolution blocks with different channel numbers in various CNN architectures. However, manual tuning via cross-validation will cost a lot of computing resources. Group convolutions have been successfully adopted to improve CNN architectures~\cite{Zhang_2017_ICCV,XieGDTH17,Ioannou_2017_CVPR}, where high-dimensional (low-dimensional) channels involve long range (short range) convolutions given the fixed number of groups. Sharing the similar philosophy, it is reasonable that the coverage of interaction (i.e., kernel size $k$ of $1D$ convolution) is proportional to channel dimension $C$. In other words, there may exist a mapping $\phi$ between $k$ and $C$: 
\begin{align}\label{fun_kp1}
C = \phi(k).
\end{align}	
The simplest mapping is a linear function, i.e., $\phi(k) = \gamma\ast k-b$. However, the relations characterized by linear function are too limited. On the other hand, it is well known that channel dimension $C$ (i.e., number of filters) usually is set to power of 2. Therefore, we introduce a possible solution by extending the  linear function $\phi(k) = \gamma\ast k-b$ to a non-linear one, i.e.,
\begin{align}\label{fun_kp2}
C = \phi(k) = 2^{(\gamma\ast k-b)}.
\end{align}	   
Then, given channel dimension $C$, kernel size $k$ can be adaptively determined by
\begin{align}\label{fun_kp3}
k = \psi(C) = \bigg|\frac{log_{2}(C)}{\gamma} + \frac{b}{\gamma}\bigg|_{odd},
\end{align}	
where $ |t|_{odd}$ indicates the nearest odd number of $t$. In this paper, we set $\gamma$ and $b$ to 2 and 1 throughout all the experiments, respectively. Clearly, through the mapping $\psi$, high-dimensional channels have longer range interaction while low-dimensional ones undergo shorter range interaction by using a non-linear mapping.

\subsection{ECA Module for Deep CNNs}
Figure~\ref{fig:method} illustrates the overview of our ECA module. After aggregating convolution features using GAP without dimensionality reduction,  ECA module first adaptively determines kernel size $k$, and then performs $1D$ convolution followed by a Sigmoid function to learn channel attention. For applying our ECA to deep CNNs, we replace SE block by our ECA module following the same configuration in~\cite{SENet18}. The resulting networks are named by ECA-Net. Figure~\ref{fig:code} gives PyTorch code of our ECA.

\section{Experiments}
In this section, we evaluate the proposed method on large-scale image classification, object detection and instance segmentation using ImageNet~\cite{imagenet_cvpr09} and MS COCO~\cite{lin2014microsoft}, respectively. Specifically, we first assess the effect of kernel size on our ECA module, and compare with state-of-the-art counterparts on ImageNet. Then, we verify the effectiveness of our ECA-Net on MS COCO using Faster R-CNN~\cite{DBLP:journals/pami/RenHG017}, Mask R-CNN~\cite{DBLP:conf/iccv/HeGDG17} and RetinaNet~\cite{DBLP:conf/iccv/LinGGHD17}. 

\subsection{Implementation Details}
To evaluate our ECA-Net on ImageNet classification, we employ four widely used CNNs as backbone models, including ResNet-50~\cite{He_2016_CVPR}, ResNet-101~\cite{He_2016_CVPR}, ResNet-512~\cite{He_2016_CVPR} and MobileNetV2~\cite{DBLP:conf/cvpr/SandlerHZZC18}. For training ResNets with our ECA, we adopt exactly the same data augmentation and hyper-parameter settings in~\cite{He_2016_CVPR,SENet18}. Specifically, the input images are randomly cropped to 224$\times$224 with random horizontal flipping. The parameters of networks are optimized by stochastic gradient descent (SGD) with weight decay of 1e-4, momentum of 0.9 and mini-batch size of 256. All models are trained within 100 epochs by setting the initial learning rate to 0.1, which is decreased by a factor of 10 per 30 epochs. For training MobileNetV2 with our ECA, we follow the settings in~\cite{DBLP:conf/cvpr/SandlerHZZC18}, where networks are trained within 400 epochs using SGD with weight decay of 4e-5, momentum of 0.9 and mini-batch size of 96. The initial learning rate is set to 0.045, and is decreased by a linear decay rate of 0.98. For testing on the validation set, the shorter side of an input image is first resized to 256 and a center crop of 224 $\times$ 224 is used for evaluation. All models are implemented by \href{https://github.com/pytorch/pytorch}{PyTorch} toolkit\footnote{\url{https://github.com/BangguWu/ECANet}}.

\begin{table*}[t]
	\centering
	\footnotesize
	\renewcommand{\arraystretch}{1.2}
	\begin{tabular}{l|c|c|c|c|c|c|c}
		\hline
		Method & Backbone Models & \#.Param. & FLOPs & Training & Inference & Top-1 & Top-5\\
		\hline
		ResNet~\cite{He_2016_CVPR} & \multirow{8}*{ResNet-50} & \textbf{24.37M} & \textbf{3.86G} & \textbf{1024 FPS} & \textbf{1855 FPS} & 75.20 & 92.52 \\
		SENet~\cite{SENet18} & ~ & 26.77M & 3.87G & 759 FPS & 1620 FPS  & 76.71 & 93.38 \\
		CBAM~\cite{Woo_2018_ECCV} & ~ & 26.77M & 3.87G & 472 FPS & 1213 FPS & 77.34 & 93.69 \\
		$A^2$-Nets~\cite{A2NIPS18}$^{\dag}$ & ~ & 33.00M  & 6.50G & N/A & N/A & 77.00 & 93.50 \\
		GCNet~\cite{Cao_2019_ICCV_Workshops} & ~ & 28.08M  & 3.87G & N/A & N/A & 77.70  & 93.66 \\
		GSoP-Net1~\cite{Gao_2019_CVPR} & ~ & 28.05M & 6.18G & 596 FPS & 1383 FPS & 77.68 & \textbf{93.98} \\
		AA-Net~\cite{1904.09925}$^{\dag,\lozenge}$ & ~ & 25.80M & 4.15G & N/A & N/A & \textbf{77.70} & 93.80 \\
		ECA-Net (Ours) & ~ & \textbf{24.37M} & \textbf{3.86G} & 785 FPS & 1805 FPS & 77.48 & 93.68 \\
		\hline
		ResNet~\cite{He_2016_CVPR} & \multirow{5}*{ResNet-101} & \textbf{42.49M} & \textbf{7.34G} & \textbf{386 FPS}  & \textbf{1174 FPS} & 76.83 & 93.48 \\
		SENet~\cite{SENet18} & ~ & 47.01M & 7.35G & 367 FPS & 1044 FPS & 77.62 & 93.93 \\
		CBAM~\cite{Woo_2018_ECCV} & ~ & 47.01M & 7.35G & 270 FPS & 635 FPS & 78.49 & 94.31 \\
		AA-Net~\cite{1904.09925}$^{\dag,\lozenge}$ & ~ & 45.40M & 8.05G & N/A & N/A & \textbf{78.70} & \textbf{94.40} \\
		ECA-Net (Ours) & ~ & \textbf{42.49M} & 7.35G & 380 FPS & 1089 FPS & 78.65 & 94.34 \\
		\hline
		ResNet~\cite{He_2016_CVPR} & \multirow{3}*{ResNet-152} & \textbf{57.40M} & \textbf{10.82G} & \textbf{281 FPS}  & \textbf{815 FPS} & 77.58 & 93.66 \\
		SENet~\cite{SENet18} & ~ & 63.68M & 10.85G & 268 FPS & 761 FPS & 78.43 & 94.27 \\
		ECA-Net (Ours) & ~ & \textbf{57.40M} & 10.83G & 279 FPS & 785 FPS & \textbf{78.92} & \textbf{94.55} \\
		\hline
		MobileNetV2~\cite{DBLP:conf/cvpr/SandlerHZZC18} & \multirow{3}*{MobileNetV2} & \textbf{3.34M} & \textbf{319.4M} & \textbf{711 FPS} & \textbf{2086 FPS} & 71.64 & 90.20 \\
		SENet &  & 3.40M & 320.1M & 671 FPS & 2000 FPS & 72.42 & 90.67 \\
		ECA-Net (Ours)& ~ & \textbf{3.34M} & 319.9M & 676 FPS & 2010 FPS & \textbf{72.56} & \textbf{90.81} \\
		\hline
	\end{tabular}
	\smallskip
	\caption{Comparison of different attention methods on ImageNet in terms of network parameters (\#.Param.), floating point operations per second (FLOPs), training or inference speed (frame per second, FPS), and Top-1/Top-5 accuracy (in \%). $\dag$: Since the source code and models of $A^2$-Nets and AA-Net are publicly unavailable, we do not compare their running time. $\lozenge$: AA-Net is trained with Inception data augmentation and different setting of learning rates.}
	\label{table2}
\end{table*}

We further evaluate our method on MS COCO using Faster R-CNN~\cite{DBLP:journals/pami/RenHG017}, Mask R-CNN~\cite{DBLP:conf/iccv/HeGDG17} and RetinaNet~\cite{DBLP:conf/iccv/LinGGHD17}, where ResNet-50 and ResNet-101 along with FPN~\cite{DBLP:conf/cvpr/LinDGHHB17} are used as backbone models. We implement all detectors by using MMDetection toolkit~\cite{mmdetection} and employ the default settings. Specifically, the shorter side of input images are resized to 800, then all models are optimized using SGD with weight decay of 1e-4, momentum of 0.9 and  mini-batch size of 8 (4 GPUs with 2 images per GPU). The learning rate is initialized to 0.01 and is decreased by a factor of 10 after 8 and 11 epochs, respectively. We train all detectors within 12 epochs on train2017 of COCO and report the results on val2017 for comparison. All programs run on a PC equipped with four RTX 2080Ti GPUs and an Intel(R) Xeon Silver 4112 CPU@2.60GHz.

\subsection{Image Classification on ImageNet-1K}

Here, we first assess the effect of kernel size on our ECA module and verify the effectiveness of our method to adaptively determine kernel size, then we compare with state-of-the-art counterparts and CNN models using ResNet-50, ResNet-101, ResNet-152 and MobileNetV2.

\subsubsection{Effect of Kernel Size ($k$) on ECA Module}
\begin{figure}[h]
	\centering
	\includegraphics[width=0.71\columnwidth]{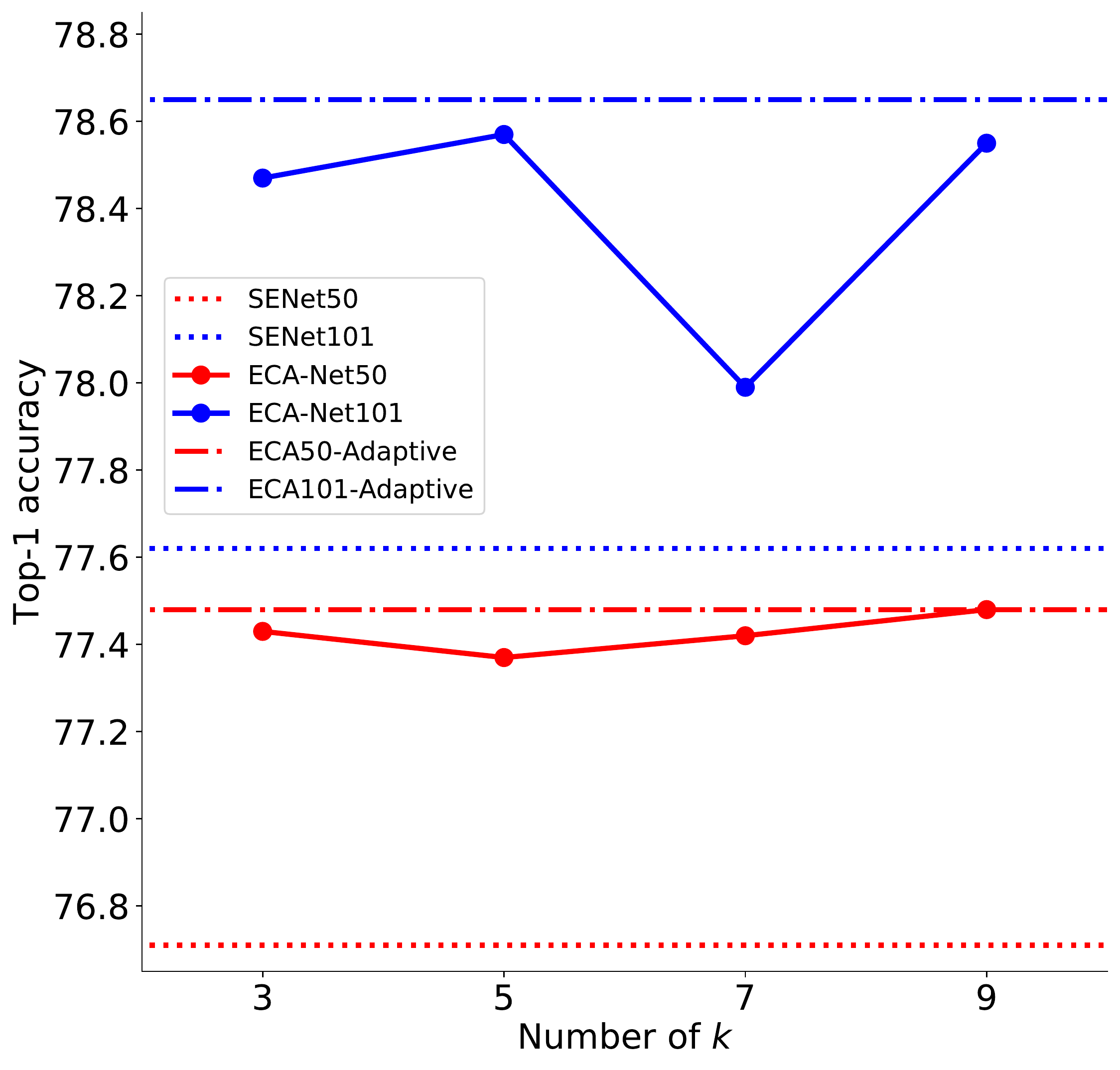} 
	\caption{Results of our ECA module with various numbers of $k$ using ResNet-50 and ResNet-101 as backbone models. Here, we also give the results of ECA module with adaptive selection of kernel size and compare with SENet as baseline.}
	\label{fig4}
\end{figure}

As shown in Eq.~(\ref{fun_eca3}), our ECA module involves a parameter $k$, i.e., kernel size of $1D$ convolution. In this part, we evaluate its effect on our ECA module and validate the effectiveness of our method for adaptive selection of kernel size. To this end, we employ ResNet-50 and ResNet-101 as backbone models, and train them with our ECA module by setting $k$ be from $3$ to $9$. The results are illustrated in Figure~\ref{fig4}, from it we have the following observations. 

Firstly, when $k$ is fixed in all convolution blocks, ECA module obtains the best results at $k=9$ and $k=5$ for ResNet-50 and ResNet-101, respectively. Since ResNet-101 has more intermediate layers that dominate performance of ResNet-101, it may prefer to small kernel size. Besides, these results show that different deep CNNs have various optimal $k$, and $k$ has a clear effect on performance of ECA-Net. Furthermore, accuracy fluctuations ($\sim$0.5\%) of ResNet-101 are larger than those ($\sim$0.15\%) of ResNet-50, and we conjecture the reason is that the deeper networks are more sensitive to the fixed kernel size than the shallower ones. Additionally, kernel size that is adaptively determined by Eq.~(\ref{fun_kp3}) usually outperforms the fixed ones, while it can avoid manual tuning of parameter $k$ via cross-validation. Above results demonstrate the effectiveness of our adaptive kernel size selection in attaining better and stable results. Finally, ECA module with various numbers of $k$ consistently outperform SE block, verifying that avoiding dimensionality reduction and local cross-channel interaction have positive effects on learning channel attention.

\subsubsection{Comparisons Using Different Deep CNNs}
\textbf{ResNet-50} We compare our ECA module with several state-of-the-art attention methods using ResNet-50 on ImageNet, including SENet~\cite{SENet18}, CBAM~\cite{Woo_2018_ECCV}, $A^2$-Nets~\cite{A2NIPS18}, AA-Net~\cite{1904.09925}, GSoP-Net1~\cite{Gao_2019_CVPR} and GCNet~\cite{Cao_2019_ICCV_Workshops}. The evaluation metrics include both efficiency (i.e., network parameters, floating point operations per second (FLOPs) and training/inference speed) and effectiveness (i.e., Top-1/Top-5 accuracy). For comparison, we duplicate the results of ResNet and SENet from ~\cite{SENet18}, and report the results of other compared methods in their original papers. To test training/inference speed of various models, we employ publicly available models of the compared CNNs, and run them on the same computing platform. The results are given in Table~\ref{table2}, where we can see that our ECA-Net shares almost the same model complexity (i.e., network parameters, FLOPs and speed) with the original ResNet-50, while achieving 2.28\% gains in Top-1 accuracy. Comparing with state-of-the-art counterparts (i.e., SENet, CBAM, $A^2$-Nets, AA-Net, GSoP-Net1 and GCNet), ECA-Net obtains better or competitive results while benefiting lower model complexity. 

\noindent\textbf{ResNet-101} Using ResNet-101 as backbone model, we compare our ECA-Net with SENet~\cite{SENet18}, CBAM~\cite{Woo_2018_ECCV} and AA-Net~\cite{1904.09925}. From Table~\ref{table2} we can see that ECA-Net outperforms the original ResNet-101 by 1.8\% with almost the same model complexity. Sharing the same tendency on ResNet-50, ECA-Net is superior to SENet and CBAM while it is very competitive to AA-Net with lower model complexity. Note that AA-Net is trained with Inception data augmentation and different setting of learning rates.

\noindent\textbf{ResNet-152} Using ResNet-152 as backbone model, we compare our ECA-Net with SENet~\cite{SENet18}. From Table~\ref{table2} we can see that ECA-Net improves the original ResNet-152 over about 1.3\% in terms of Top-1 accuracy with almost the same model complexity. Comparing with SENet, ECA-Net achieves 0.5\% gain in terms of Top-1 with lower model complexity. The results with respect to ResNet-50, ResNet-101 and ResNet-152 demonstrate the effectiveness of our ECA module on the widely used ResNet architectures.

\noindent\textbf{MobileNetV2} Besides ResNet architectures, we also verify the effectiveness of our ECA module on lightweight CNN architectures. To this end, we employ MobileNetV2~\cite{DBLP:conf/cvpr/SandlerHZZC18} as backbone model and compare our ECA module with SE block. In particular, we integrate SE block and ECA module in convolution layer before residual connection lying in each 'bottleneck' of MobileNetV2, and parameter $r$ of SE block is set to 8. All models are trained using exactly the same settings. The results in Table~\ref{table2} show our ECA-Net improves the original MobileNetV2 and SENet by about 0.9\% and 0.14\% in terms of Top-1 accuracy, respectively. Furthermore, our ECA-Net has smaller model size and faster training/inference speed than SENet. Above results verify the efficiency and effectiveness of our ECA module again.

\begin{table}[h]
	\centering
	\footnotesize
	\renewcommand{\arraystretch}{1.2}
	\begin{tabular}{l|c|c|c|c}
		\hline
		CNN Models & \#.Param. & FLOPs & Top-1 & Top-5\\
		\hline
		ResNet-200 & 74.45M & 14.10G & 78.20 & 94.00 \\
		Inception-v3  & 25.90M & 5.36G & 77.45 & 93.56 \\
		ResNeXt-101 & 46.66M & 7.53G & 78.80 & 94.40 \\
		DenseNet-264 (k=32) & 31.79M & 5.52G & 77.85 & 93.78 \\
		DenseNet-161 (k=48) & 27.35M & 7.34G & 77.65 & 93.80 \\
		\hline
		ECA-Net50 (Ours) & 24.37M & 3.86G & 77.48 & 93.68 \\
		ECA-Net101 (Ours) & 42.49M & 7.35G & 78.65 & 94.34 \\
		\hline
	\end{tabular}
	\smallskip
	\caption{Comparisons with state-of-the-art CNNs on ImageNet.}
	\label{table3:SOTA}
\end{table}

\subsubsection{Comparisons with Other CNN Models}
At the end of this part, we compare our ECA-Net50 and ECA-Net101 with other state-of-the-art CNN models, including ResNet-200~\cite{DBLP:conf/eccv/HeZRS16}, Inception-v3~\cite{DBLP:conf/cvpr/SzegedyVISW16}, ResNeXt~\cite{XieGDTH17}, DenseNet~\cite{Huang_2017_CVPR}. These CNN models have deeper and wider architectures, and their results all are copied from the original papers. As presented in Table~\ref{table3:SOTA}, ECA-Net101 outperforms ResNet-200, indicating that our ECA-Net can improve the performance of deep CNNs using much less computational cost. Meanwhile, our ECA-Net101 is very competitive to ResNeXt-101, while the latter one employs more convolution filters and expensive group convolutions. In addition, ECA-Net50 is comparable to DenseNet-264 (k=32), DenseNet-161 (k=48) and Inception-v3, but it has lower model complexity. All above results demonstrate that our ECA-Net performs favorably against state-of-the-art CNNs while benefiting much lower model complexity. Note that our ECA also has great potential to further improve the performance of the compared CNN models.

\begin{table*}[t]
	\centering
	\footnotesize
	\renewcommand{\arraystretch}{1.2}
	\begin{tabular}{l|c|c|c|c|c|c|c|c|c}
		\hline
		Methods & Detectors & \#.Param. & GFLOPs & AP & $AP_{50}$ & $AP_{75}$ & $AP_S$ & $AP_M$ & $AP_L$\\
		\hline
		ResNet-50 &  \multirow{6}*{Faster R-CNN} & \textbf{41.53 M} & \textbf{207.07}  & 36.4 & 58.2 & 39.2 & 21.8 & 40.0 & 46.2 \\
		+ SE block & & 44.02 M & 207.18 & 37.7 & 60.1 & 40.9 & 22.9 & 41.9 & \textbf{48.2} \\
		+ ECA (Ours) & & \textbf{41.53 M} & 207.18 & \textbf{38.0} & \textbf{60.6} & \textbf{40.9} & \textbf{23.4} & \textbf{42.1} & 48.0 \\
		\cline{1-1}\cline{3-10}
		ResNet-101 & & \textbf{60.52 M} & \textbf{283.14} & 38.7 & 60.6 & 41.9 & 22.7 & 43.2 & 50.4 \\
		+ SE block & & 65.24 M & 283.33 & 39.6 & 62.0 & 43.1 & 23.7 & 44.0 & \textbf{51.4} \\
		+ ECA (Ours) & & \textbf{60.52 M} & 283.32 & \textbf{40.3} & \textbf{62.9} & \textbf{44.0} & \textbf{24.5} & \textbf{44.7} & 51.3 \\
		\hline
		ResNet-50 &  \multirow{8}*{Mask R-CNN} & \textbf{44.18 M} & \textbf{275.58} & 37.2 & 58.9 & 40.3 & 22.2 & 40.7 & 48.0 \\
		+ SE block & & 46.67 M & 275.69 & 38.7 & 60.9 & 42.1 & 23.4 & 42.7 & \textbf{50.0} \\
		+ 1 NL & & 46.50 M & 288.70 & 38.0  & 59.8  & 41.0 & N/A & N/A  & N/A  \\
		+ GC block & & 46.90 M & 279.60 & \textbf{39.4} &  \textbf{61.6}  & \textbf{42.4} & N/A & N/A & N/A \\
		+ ECA (Ours) & & \textbf{44.18 M} & 275.69 & 39.0 & 61.3 & 42.1 & \textbf{24.2} & \textbf{42.8} & 49.9 \\
		\cline{1-1}\cline{3-10}
		ResNet-101 & & \textbf{63.17 M} & \textbf{351.65} & 39.4 &60.9 & 43.3 & 23.0 & 43.7 & 51.4 \\
		+ SE block & & 67.89 M & 351.84 & 40.7 & 62.5 & 44.3 & 23.9 & 45.2 & 52.8 \\
		+ ECA (Ours) & & \textbf{63.17 M} & 351.83 & \textbf{41.3} & \textbf{63.1} & \textbf{44.8} & \textbf{25.1} & \textbf{45.8} & \textbf{52.9} \\
		\hline
		ResNet-50 &  \multirow{6}*{RetinaNet} & \textbf{37.74 M} & \textbf{239.32} & 35.6 & 55.5 & 38.2 & 20.0 & 39.6 & 46.8 \\
		+ SE block & & 40.23 M & 239.43 & 37.1 & 57.2 & \textbf{39.9} & 21.2 & 40.7 & \textbf{49.3} \\
		+ ECA (Ours) & & \textbf{37.74 M} & 239.43 & \textbf{37.3} & \textbf{57.7} & 39.6 & \textbf{21.9} & \textbf{41.3} & 48.9 \\
		\cline{1-1}\cline{3-10}
		ResNet-101 & & \textbf{56.74 M} & \textbf{315.39} & 37.7 & 57.5 & 40.4 & 21.1 & 42.2 & 49.5 \\
		+ SE block & & 61.45 M & 315.58 & 38.7 & 59.1 & 41.6 & 22.1 & 43.1 & \textbf{50.9} \\
		+ ECA (Ours) & & \textbf{56.74 M} & 315.57 & \textbf{39.1} & \textbf{59.9} & \textbf{41.8} & \textbf{22.8} &  \textbf{43.4} & 50.6 \\
		\hline
	\end{tabular}
	\smallskip
	\caption{Object detection results of different methods on COCO val2017.}
	\label{table4:detection}
\end{table*}

\subsection{Object Detection on MS COCO}
In this subsection, we evaluate our ECA-Net on object detection task using Faster R-CNN~\cite{DBLP:journals/pami/RenHG017}, Mask R-CNN~\cite{DBLP:conf/iccv/HeGDG17} and RetinaNet~\cite{DBLP:conf/iccv/LinGGHD17}. We mainly compare ECA-Net with ResNet and SENet. All CNN models are pre-trained on ImageNet, then are transferred to MS COCO by fine-tuning.  

\subsubsection{Comparisons Using Faster R-CNN}
Using Faster R-CNN as the basic detector, we employ ResNets of 50 and 101 layers along with FPN~\cite{DBLP:conf/cvpr/LinDGHHB17} as backbone models. As shown in Table~\ref{table4:detection}, integration of either SE block or our ECA module can improve performance of object detection by a clear margin. Meanwhile, our ECA outperforms SE block by 0.3\% and 0.7\% in terms of AP using ResNet-50 and ResNet-101, respectively. 

\vspace{-0.1cm}
\subsubsection{Comparisons Using Mask R-CNN}
We further exploit Mask R-CNN to verify the effectiveness of our ECA-Net on object detection task.  As shown in Table~\ref{table4:detection}, our ECA module is superior to the original ResNet by 1.8\% and 1.9\% in terms of AP under the settings of 50 and 101 layers, respectively. Meanwhile, ECA module achieves 0.3\% and 0.6\% gains over SE block using ResNet-50 and ResNet-101 as backbone models, respectively. Using ResNet-50, ECA is superior to one NL~\cite{Wang_2018_CVPR}, and is comparable to GC block~\cite{Cao_2019_ICCV_Workshops} using lower model complexity.

\vspace{-0.3cm}
\subsubsection{Comparisons Using RetinaNet}
Additionally, we verify the effectiveness of our ECA-Net on object detection using one-stage detector, i.e., RetinaNet.  As compared in Table~\ref{table4:detection}, our ECA-Net outperforms the original ResNet by 1.8\% and 1.4\% in terms of AP for the networks of 50 and 101 layers, respectively. Meanwhile, ECA-Net improves SE-Net over 0.2\% and 0.4\%  for ResNet-50 and ResNet-101, respectively. In summary, the results in Table~\ref{table4:detection} demonstrate that our ECA-Net can well generalize to object detection task. Specifically, ECA module brings clear improvement over the original ResNet, while outperforming  SE block using lower model complexity. In particular, our ECA module achieves more gains for small objects, which are usually more difficult to be detected.  

\begin{table}[t]
	\centering
	\footnotesize
	\renewcommand{\arraystretch}{1.2}
	\begin{tabular}{l|c|c|c|c|c|c}
		\hline
		Methods  & AP & $AP_{50}$ & $AP_{75}$ & $AP_S$ & $AP_M$ & $AP_L$\\
		\hline
		ResNet-50  & 34.1 & 55.5 & 36.2 & 16.1 & 36.7 & 50.0 \\
		+ SE block & 35.4 & 57.4 & \textbf{37.8} & 17.1 & 38.6 & 51.8 \\
		+ 1 NL  & 34.7 & 56.7  & 36.6 & N/A & N/A & N/A \\
		+ GC block & \textbf{35.7}   & \textbf{58.4}  & 37.6 & N/A & N/A & N/A \\
		+ ECA (Ours) & 35.6 & 58.1 & 37.7 & \textbf{17.6} & \textbf{39.0} & \textbf{51.8} \\
		\hline
		ResNet-101 & 35.9 & 57.7 & 38.4 & 16.8 & 39.1 & 53.6 \\
		+ SE block  & 36.8 & 59.3 & 39.2 & 17.2 & 40.3 & 53.6 \\
		+ ECA (Ours) & \textbf{37.4} & \textbf{59.9} & \textbf{39.8} & \textbf{18.1} & \textbf{41.1} & \textbf{54.1} \\
		\hline
	\end{tabular}
	\smallskip
	\caption{Instance segmentation results of different methods using Mask R-CNN on COCO val2017.}
	\label{table:instance}
\end{table}

\subsection{Instance Segmentation on MS COCO}	
Then, we give instance segmentation results of our ECA module using Mask R-CNN on MS COCO. As compared in Table~\ref{table:instance}, ECA module achieves notable gain over the original ResNet while performing better than SE block with less model complexity. For ResNet-50 as backbone, ECA with lower model complexity is superior one NL~\cite{Wang_2018_CVPR}, and is comparable to GC block~\cite{Cao_2019_ICCV_Workshops}. These results verify our ECA module has good generalization ability for various tasks. 

\section{Conclusion}
In this paper, we focus on learning effective channel attention for deep CNNs with low model complexity. To this end, we propose an efficient channel attention (ECA) module, which generates channel attention through a fast $1D$ convolution, whose kernel size can be adaptively determined by a non-linear mapping of channel dimension. Experimental results demonstrate our ECA is an extremely lightweight plug-and-play block to improve the performance of various deep CNN architectures, including the  widely used ResNets and lightweight MobileNetV2. Moreover, our ECA-Net exhibits good generalization ability in object detection and instance segmentation tasks. In future, we will apply our ECA module to more CNN architectures (e.g., ResNeXt and Inception~\cite{DBLP:conf/cvpr/SzegedyVISW16}) and further investigate incorporation of ECA with spatial attention module.

{\small
	\bibliographystyle{ieee_fullname}
	\bibliography{arxiv_4422}

\begin{thebibliography}{10}\itemsep=-1pt

\bibitem{1904.09925}
Irwan Bello, Barret Zoph, Ashish Vaswani, Jonathon Shlens, and Quoc~V. Le.
\newblock Attention augmented convolutional networks.
\newblock {\em arXiv:1904.09925}, 2019.

\bibitem{Cao_2019_ICCV_Workshops}
Yue Cao, Jiarui Xu, Stephen Lin, Fangyun Wei, and Han Hu.
\newblock Gcnet: Non-local networks meet squeeze-excitation networks and
  beyond.
\newblock In {\em ICCV Workshops}, 2019.

\bibitem{mmdetection}
Kai Chen, Jiaqi Wang, Jiangmiao Pang, Yuhang Cao, Yu Xiong, Xiaoxiao Li,
  Shuyang Sun, Wansen Feng, Ziwei Liu, Jiarui Xu, Zheng Zhang, Dazhi Cheng,
  Chenchen Zhu, Tianheng Cheng, Qijie Zhao, Buyu Li, Xin Lu, Rui Zhu, Yue Wu,
  Jifeng Dai, Jingdong Wang, Jianping Shi, Wanli Ouyang, Chen~Change Loy, and
  Dahua Lin.
\newblock {MMDetection}: Open mmlab detection toolbox and benchmark.
\newblock {\em arXiv:1906.07155}, 2019.

\bibitem{A2NIPS18}
Yunpeng Chen, Yannis Kalantidis, Jianshu Li, Shuicheng Yan, and Jiashi Feng.
\newblock {A$^{2}$-Nets}: Double attention networks.
\newblock In {\em NIPS}, 2018.

\bibitem{DBLP:conf/cvpr/Chollet17}
Fran{\c{c}}ois Chollet.
\newblock Xception: Deep learning with depthwise separable convolutions.
\newblock In {\em CVPR}, 2017.

\bibitem{imagenet_cvpr09}
J. Deng, W. Dong, R. Socher, L.-J. Li, K. Li, and L. Fei-Fei.
\newblock {ImageNet}: A large-scale hierarchical image database.
\newblock In {\em CVPR}, 2009.

\bibitem{Fu_2019_CVPR}
Jun Fu, Jing Liu, Haijie Tian, Yong Li, Yongjun Bao, Zhiwei Fang, and Hanqing
  Lu.
\newblock Dual attention network for scene segmentation.
\newblock In {\em CVPR}, 2019.

\bibitem{DBLP:conf/nips/GaoWJ18}
Hongyang Gao, Zhengyang Wang, and Shuiwang Ji.
\newblock Channelnets: Compact and efficient convolutional neural networks via
  channel-wise convolutions.
\newblock In {\em NeurIPS}, 2018.

\bibitem{Gao_2019_CVPR}
Zilin Gao, Jiangtao Xie, Qilong Wang, and Peihua Li.
\newblock Global second-order pooling convolutional networks.
\newblock In {\em CVPR}, 2019.

\bibitem{DBLP:conf/iccv/HeGDG17}
Kaiming He, Georgia Gkioxari, Piotr Doll{\'{a}}r, and Ross~B. Girshick.
\newblock Mask {R-CNN}.
\newblock In {\em ICCV}, pages 2980--2988, 2017.

\bibitem{He_2016_CVPR}
Kaiming He, Xiangyu Zhang, Shaoqing Ren, and Jian Sun.
\newblock Deep residual learning for image recognition.
\newblock In {\em CVPR}, 2016.

\bibitem{DBLP:conf/eccv/HeZRS16}
Kaiming He, Xiangyu Zhang, Shaoqing Ren, and Jian Sun.
\newblock Identity mappings in deep residual networks.
\newblock In {\em ECCV}, 2016.

\bibitem{DBLP:conf/nips/HuSASV18}
Jie Hu, Li Shen, Samuel Albanie, Gang Sun, and Andrea Vedaldi.
\newblock Gather-excite: Exploiting feature context in convolutional neural
  networks.
\newblock In {\em NeurIPS}, 2018.

\bibitem{SENet18}
Jie Hu, Li Shen, and Gang Sun.
\newblock Squeeze-and-excitation networks.
\newblock In {\em CVPR}, 2018.

\bibitem{Huang_2017_CVPR}
Gao Huang, Zhuang Liu, Laurens van~der Maaten, and Kilian~Q. Weinberger.
\newblock Densely connected convolutional networks.
\newblock In {\em CVPR}, 2017.

\bibitem{Ioannou_2017_CVPR}
Yani Ioannou, Duncan Robertson, Roberto Cipolla, and Antonio Criminisi.
\newblock Deep roots: Improving cnn efficiency with hierarchical filter groups.
\newblock In {\em CVPR}, 2017.

\bibitem{nips2012cnn}
Alex Krizhevsky, Ilya Sutskever, and Geoffrey~E Hinton.
\newblock Image{N}et classification with deep convolutional neural networks.
\newblock In {\em NIPS}, 2012.

\bibitem{DBLP-1901-01493}
Huayu Li.
\newblock Channel locality block: {A} variant of squeeze-and-excitation.
\newblock {\em arXiv}, 1901.01493, 2019.

\bibitem{LiXWZ17}
Peihua Li, Jiangtao Xie, Qilong Wang, and Wangmeng Zuo.
\newblock Is second-order information helpful for large-scale visual
  recognition?
\newblock In {\em ICCV}, 2017.

\bibitem{LiYanghao_2017_ICCV}
Yanghao Li, Naiyan Wang, Jiaying Liu, and Xiaodi Hou.
\newblock Factorized bilinear models for image recognition.
\newblock In {\em ICCV}, 2017.

\bibitem{DBLP:conf/cvpr/LinDGHHB17}
Tsung{-}Yi Lin, Piotr Doll{\'{a}}r, Ross~B. Girshick, Kaiming He, Bharath
  Hariharan, and Serge~J. Belongie.
\newblock Feature pyramid networks for object detection.
\newblock In {\em CVPR}, pages 936--944, 2017.

\bibitem{DBLP:conf/iccv/LinGGHD17}
Tsung{-}Yi Lin, Priya Goyal, Ross~B. Girshick, Kaiming He, and Piotr
  Doll{\'{a}}r.
\newblock Focal loss for dense object detection.
\newblock In {\em ICCV}, 2017.

\bibitem{lin2014microsoft}
Tsung-Yi Lin, Michael Maire, Serge Belongie, James Hays, Pietro Perona, Deva
  Ramanan, Piotr Doll{\'a}r, and C~Lawrence Zitnick.
\newblock {Microsoft COCO}: Common objects in context.
\newblock In {\em ECCV}, 2014.

\bibitem{DBLP:conf/eccv/MaZZS18}
Ningning Ma, Xiangyu Zhang, Hai{-}Tao Zheng, and Jian Sun.
\newblock Shufflenet {V2:} {Practical} guidelines for efficient {CNN}
  architecture design.
\newblock In {\em ECCV}, 2018.

\bibitem{DBLP:conf/icml/NairH10}
Vinod Nair and Geoffrey~E. Hinton.
\newblock Rectified linear units improve restricted boltzmann machines.
\newblock In {\em ICML}, 2010.

\bibitem{DBLP:journals/pami/RenHG017}
Shaoqing Ren, Kaiming He, Ross~B. Girshick, and Jian Sun.
\newblock Faster {R-CNN:} {Towards} real-time object detection with region
  proposal networks.
\newblock {\em {IEEE} Trans. Pattern Anal. Mach. Intell.}, 39(6):1137--1149,
  2017.

\bibitem{DBLP:journals/tmi/RoyNW19}
Abhijit~Guha Roy, Nassir Navab, and Christian Wachinger.
\newblock Recalibrating fully convolutional networks with spatial and channel
  "squeeze and excitation" blocks.
\newblock {\em {IEEE} Trans. Med. Imaging}, 38(2):540--549, 2019.

\bibitem{DBLP:conf/cvpr/SandlerHZZC18}
Mark Sandler, Andrew~G. Howard, Menglong Zhu, Andrey Zhmoginov, and
  Liang{-}Chieh Chen.
\newblock Mobilenetv2: Inverted residuals and linear bottlenecks.
\newblock In {\em CVPR}, 2018.

\bibitem{Simonyan15}
K. Simonyan and A. Zisserman.
\newblock Very deep convolutional networks for large-scale image recognition.
\newblock In {\em ICLR}, 2015.

\bibitem{Szegedy_2015_CVPR}
Christian Szegedy, Wei Liu, Yangqing Jia, Pierre Sermanet, Scott Reed, Dragomir
  Anguelov, Dumitru Erhan, Vincent Vanhoucke, and Andrew Rabinovich.
\newblock Going deeper with convolutions.
\newblock In {\em CVPR}, 2015.

\bibitem{DBLP:conf/cvpr/SzegedyVISW16}
Christian Szegedy, Vincent Vanhoucke, Sergey Ioffe, Jonathon Shlens, and
  Zbigniew Wojna.
\newblock Rethinking the inception architecture for computer vision.
\newblock In {\em CVPR}, 2016.

\bibitem{Wang_2018_CVPR}
Xiaolong Wang, Ross Girshick, Abhinav Gupta, and Kaiming He.
\newblock Non-local neural networks.
\newblock In {\em CVPR}, 2018.

\bibitem{Woo_2018_ECCV}
Sanghyun Woo, Jongchan Park, Joon-Young Lee, and In~So Kweon.
\newblock {CBAM}: Convolutional block attention module.
\newblock In {\em ECCV}, 2018.

\bibitem{XieGDTH17}
Saining Xie, Ross~B. Girshick, Piotr Doll{\'{a}}r, Zhuowen Tu, and Kaiming He.
\newblock Aggregated residual transformations for deep neural networks.
\newblock In {\em CVPR}, 2017.

\bibitem{DBLP:conf/eccv/ZeilerF14}
Matthew~D. Zeiler and Rob Fergus.
\newblock Visualizing and understanding convolutional networks.
\newblock In {\em ECCV}, pages 818--833, 2014.

\bibitem{DBLP:conf/cvpr/Zhang18}
Dong{-}Qing Zhang.
\newblock Clcnet: Improving the efficiency of convolutional neural network
  using channel local convolutions.
\newblock In {\em CVPR}, 2018.

\bibitem{Zhang_2017_ICCV}
Ting Zhang, Guo-Jun Qi, Bin Xiao, and Jingdong Wang.
\newblock Interleaved group convolutions.
\newblock In {\em ICCV}, 2017.

\bibitem{DBLP:conf/cvpr/ZhangZLS18}
Xiangyu Zhang, Xinyu Zhou, Mengxiao Lin, and Jian Sun.
\newblock Shufflenet: An extremely efficient convolutional neural network for
  mobile devices.
\newblock In {\em CVPR}, 2018.

\end{thebibliography}
}

\begin{table}[t]
	\centering
	\footnotesize
	\renewcommand{\arraystretch}{1.2}
	\begin{tabular}{l|c|c|c|c|c}
		\hline
		Method & CNNs & \#.Param. & GFLOPs  & Top-1 & Top-5\\
		\hline
		ResNet~\cite{He_2016_CVPR} & \multirow{4}*{R-18} & \textbf{11.148M} & \textbf{1.699} &  70.40 & 89.45 \\
		SENet~\cite{SENet18} & ~ & 11.231M & 1.700 & 70.59 & 89.78 \\
		CBAM~\cite{Woo_2018_ECCV} & ~ & 11.234M & 1.700 & 70.73 & 89.91 \\
		ECA-Net (Ours) & ~ & \textbf{11.148M} & 1.700 & \textbf{70.78} & \textbf{89.92} \\
		\hline
		ResNet~\cite{He_2016_CVPR} & \multirow{4}*{R-34} & \textbf{20.788M} & \textbf{3.427}  & 73.31 & 91.40 \\
		SENet~\cite{SENet18} & ~ & 20.938M & 3.428 &  73.87 & 91.65 \\
		CBAM~\cite{Woo_2018_ECCV} & ~ & 20.943M & 3.428 & 74.01 & 91.76 \\
		ECA-Net (Ours) & ~ & \textbf{20.788M} & 3.428 & \textbf{74.21} & \textbf{91.83} \\
		\hline
	\end{tabular}
	\smallskip
	\caption{Comparison of different methods using ResNet-18 (R-18) and ResNet-34 (R-34) on ImageNet in terms of network parameters (\#.Param.), floating point operations per second (FLOPs), and Top-1/Top-5 accuracy (in \%).}
	\label{tableA2}
\end{table}

\section*{Appendix I: Comparison of Different Methods using ResNet-18 and ResNet-34 on ImageNet}

Here, we compare different attention methods using ResNet-18 and ResNet-34 on ImageNet. The results are listed in Table~\ref{tableA2}, where the results of ResNet, SENet and CBAM are duplicated from~\cite{Woo_2018_ECCV}, and we train ECA-Net using the settings of hyper-parameters with~\cite{Woo_2018_ECCV}. From Table~\ref{tableA2}, we can see that our ECA-Net improves the original ResNet-18 and ResNet-34 over 0.38\% abd 0.9\% in Top-1 accuracy, respectively. Comparing with SENet and CBAM, our ECA-Net achieves better performance using less model complexity, showing the effectiveness of the proposed ECA module.  

\section*{Appendix II: Stacking More 1D Convolutions in ECA Module}

Intuitively, more 1D convolutions stacked in ECA module may bring further improvement, due to increase of modeling capability. Actually, we found that one extra 1D convolution brings trivial  gains ($\sim$0.1\%) at the cost of slightly increasing complexity, but more 1D convolutions degrade performance, which may be caused by that more 1D convolutions make gradient backpropagation more difficult. Therefore, our final ECA module contains only one 1D convolution.

\begin{figure}[t]
	\centering
	\includegraphics[width=1.0\linewidth]{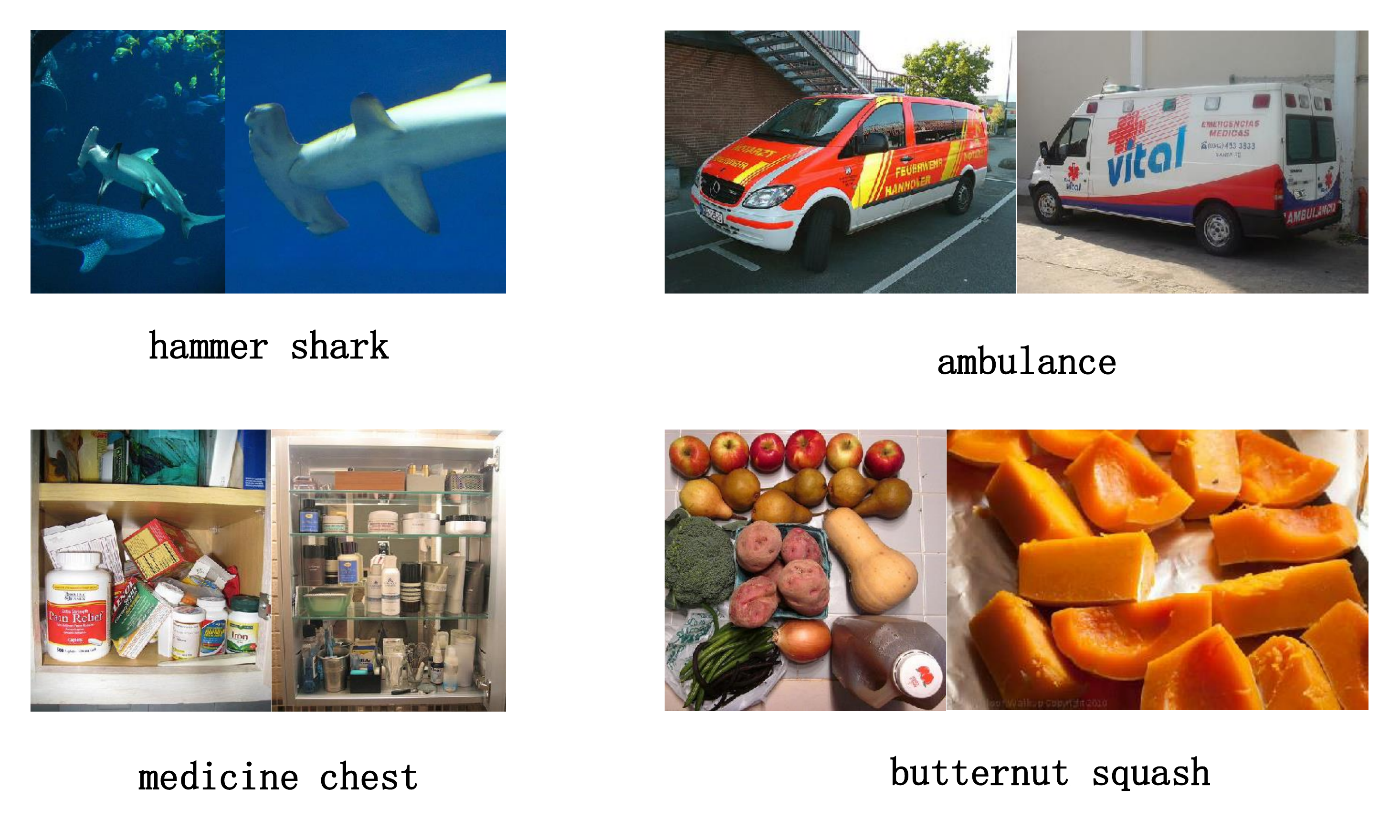}
	\caption{Example images of four random sampled classes on ImageNet, including \textit{hammerhead shark}, \textit{ambulance}, \textit{medicine chest} and \textit{butternut squash}.}
	\label{examples}
\end{figure}

\begin{figure*}
	\subfigure[conv\_2\_1]{
		\begin{minipage}[t]{0.25\linewidth}
			\centering
			\includegraphics[width=0.8\columnwidth]{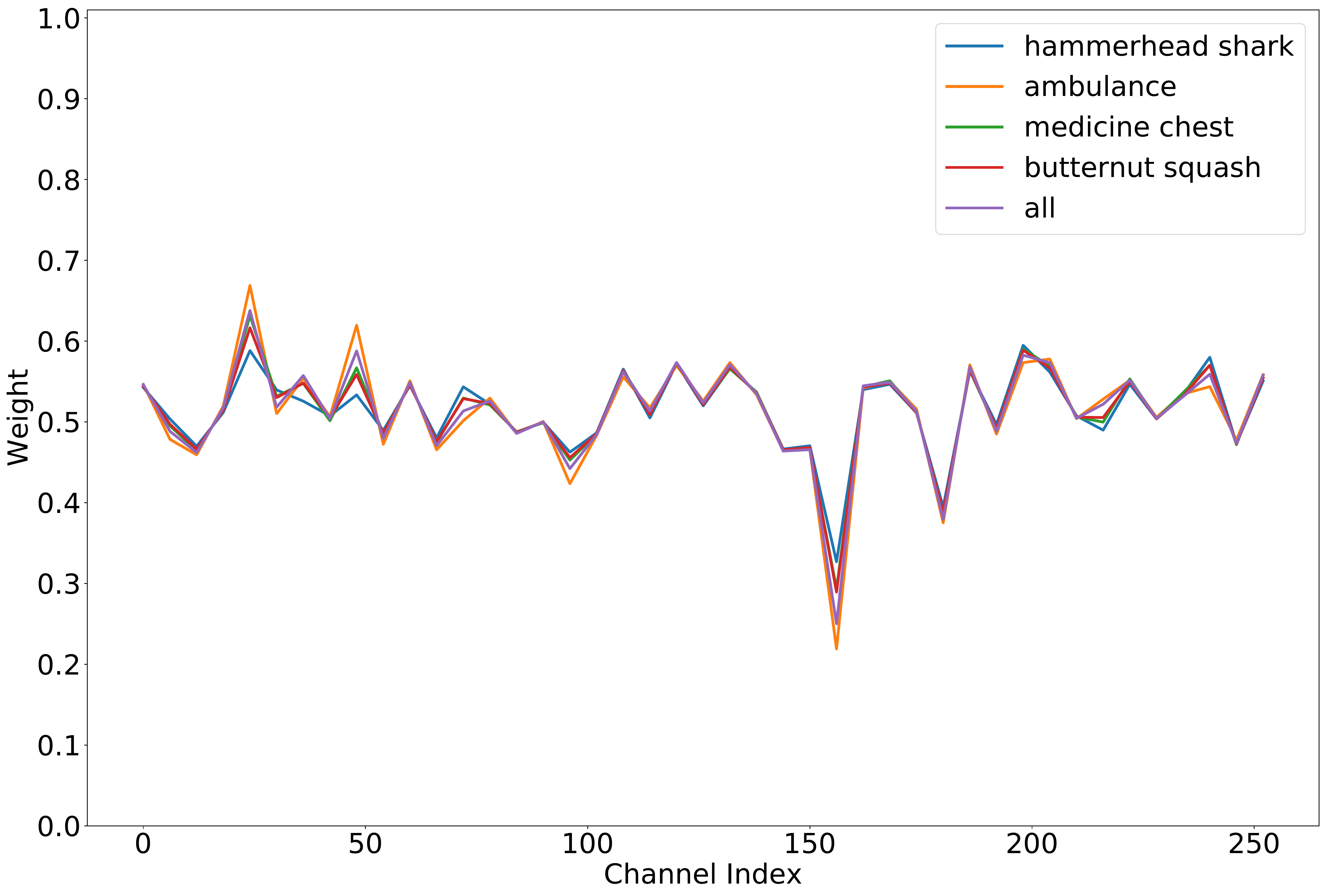}
			\includegraphics[width=0.8\columnwidth]{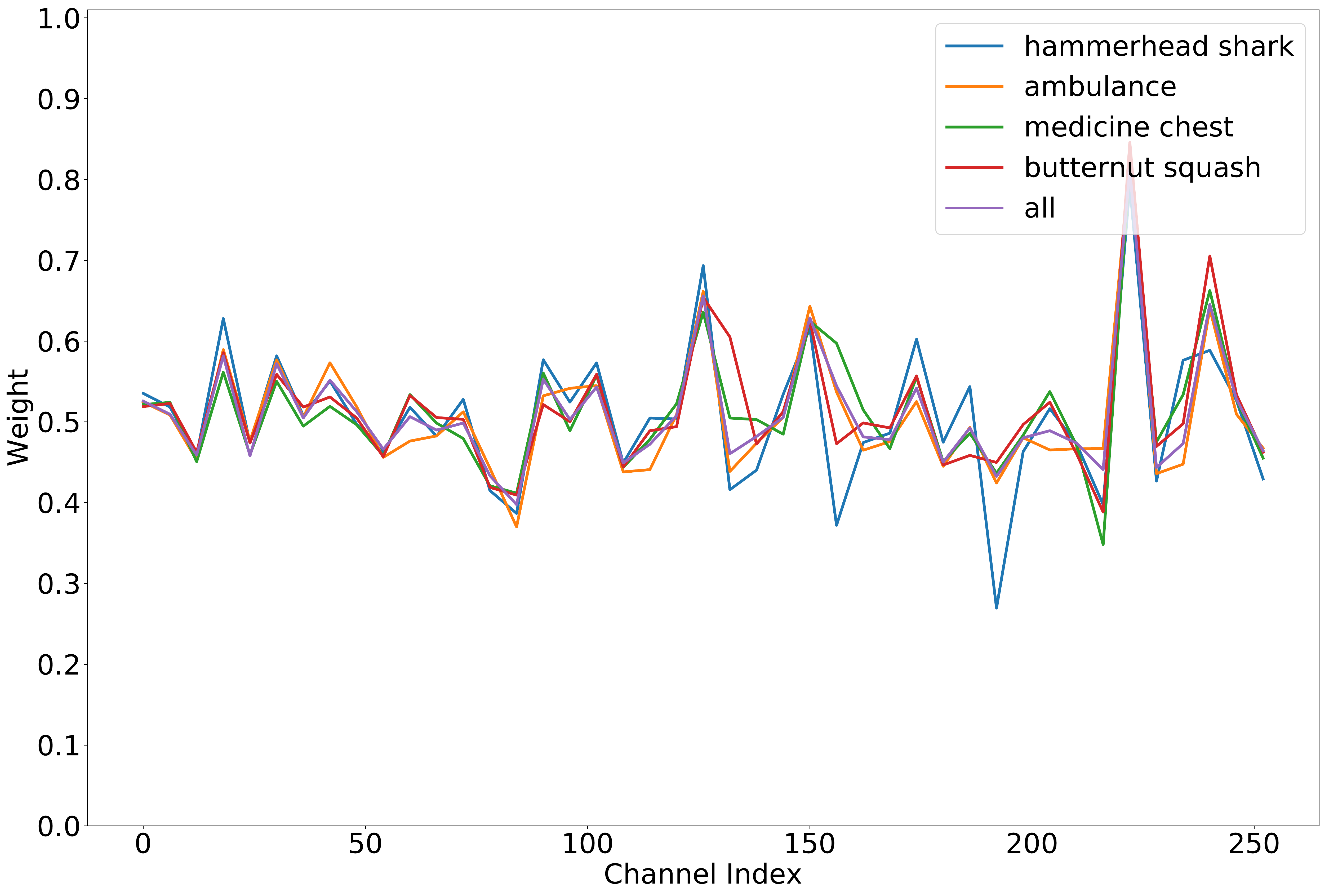}
		\end{minipage}%
	}%
	\subfigure[conv\_2\_2]{
		\begin{minipage}[t]{0.25\linewidth}
			\centering
			\includegraphics[width=0.8\columnwidth]{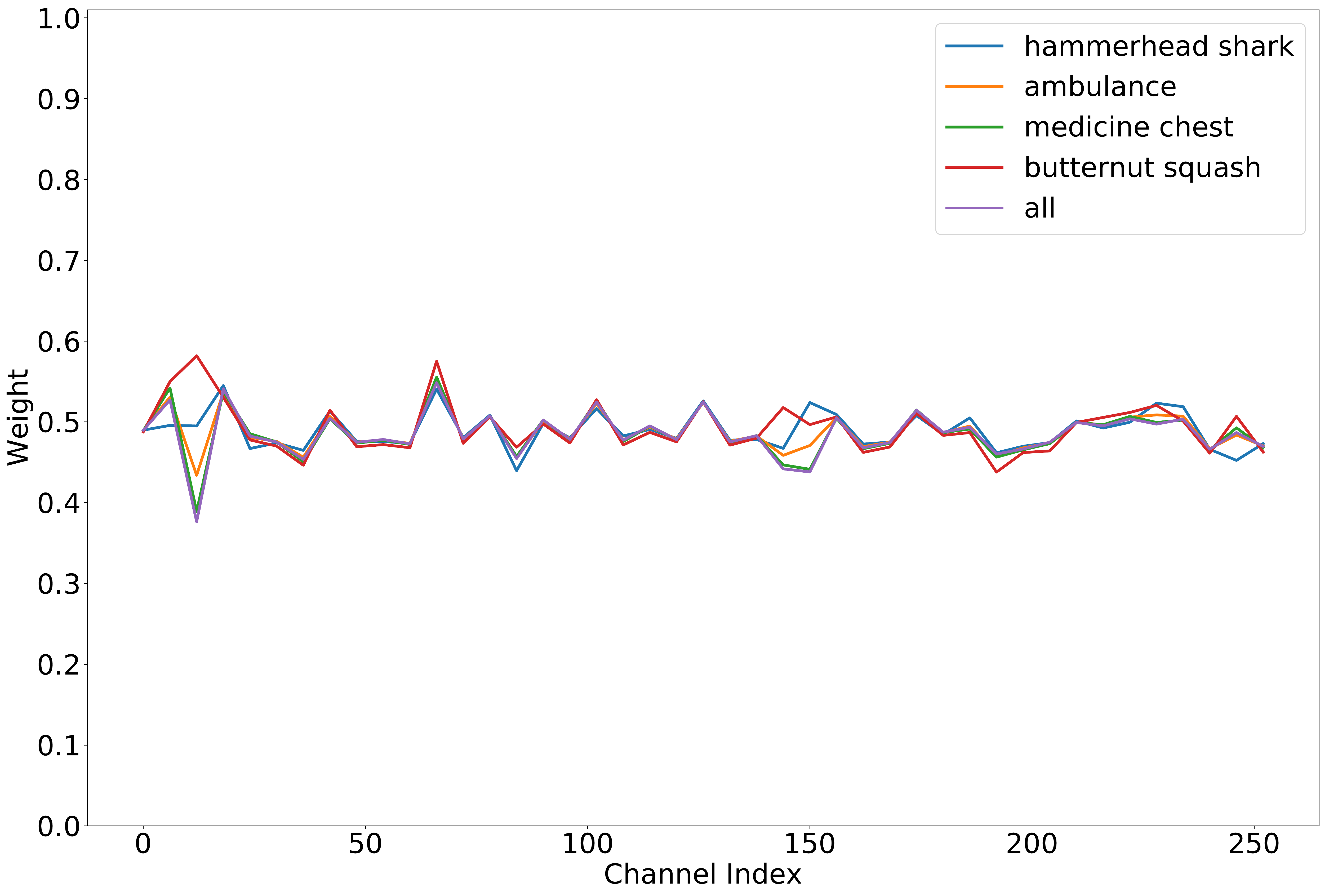}						\includegraphics[width=0.8\columnwidth]{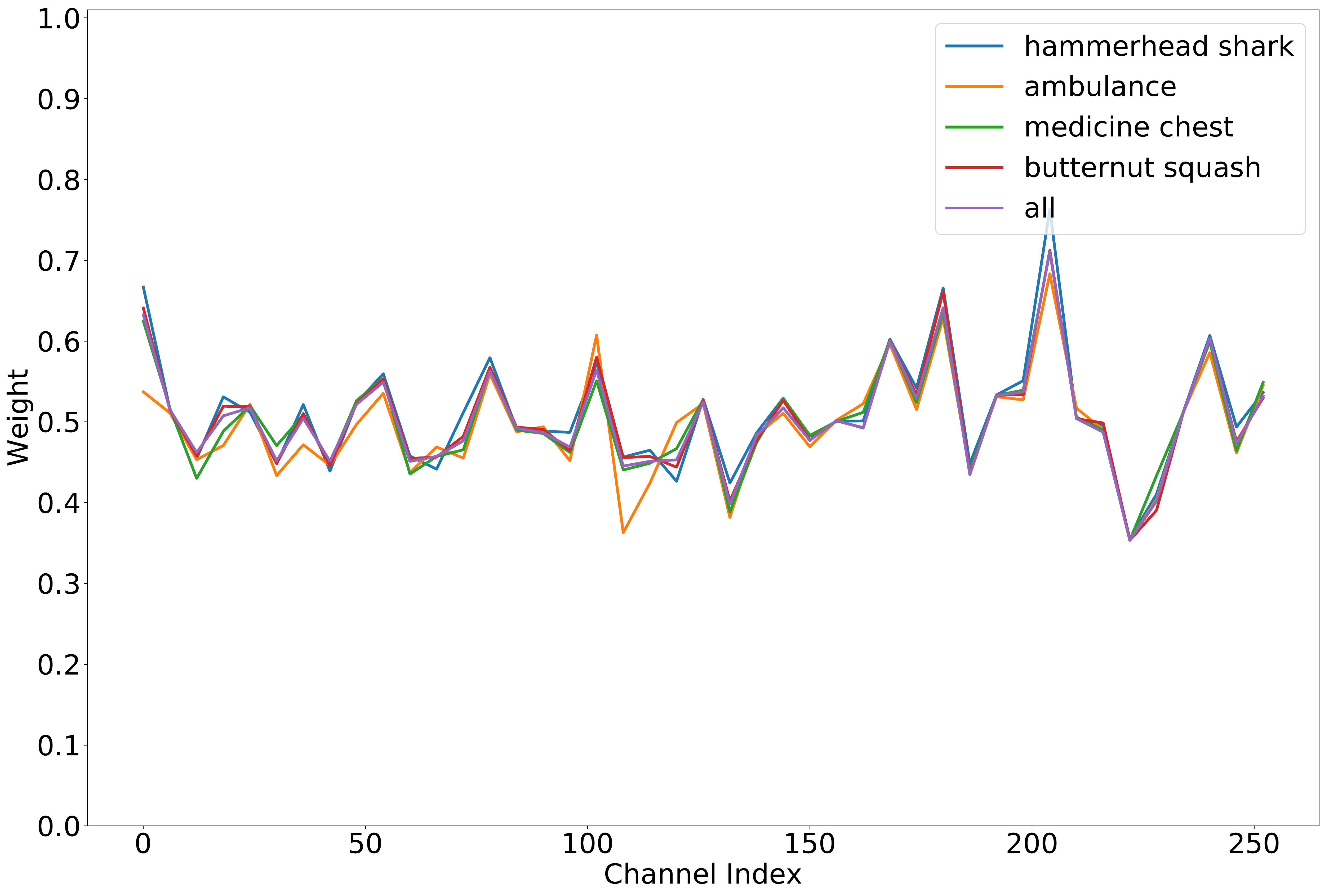}
		\end{minipage}%
	}%
	\subfigure[conv\_2\_3]{
		\begin{minipage}[t]{0.25\linewidth}
			\centering
			\includegraphics[width=0.8\columnwidth]{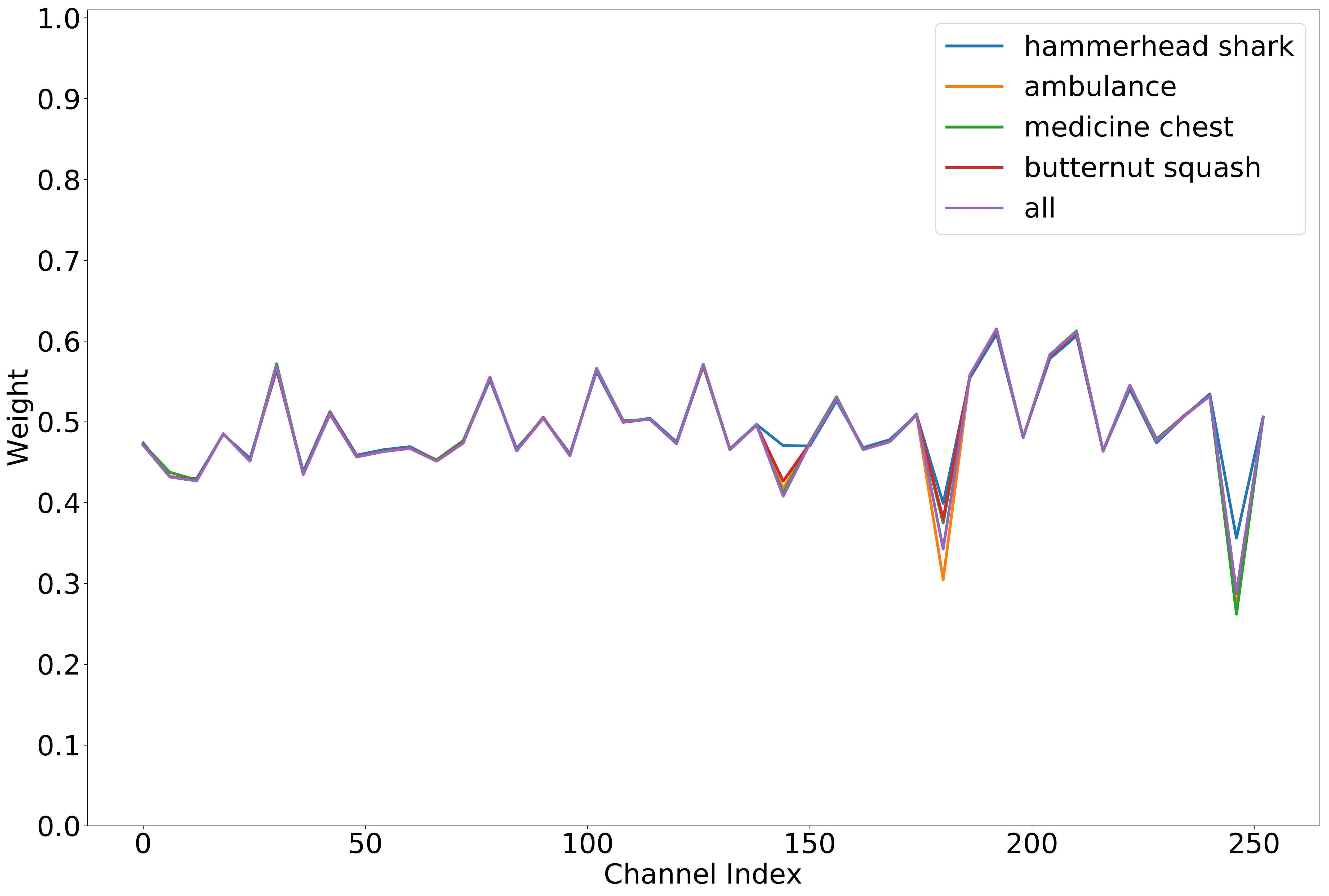}
			\includegraphics[width=0.8\columnwidth]{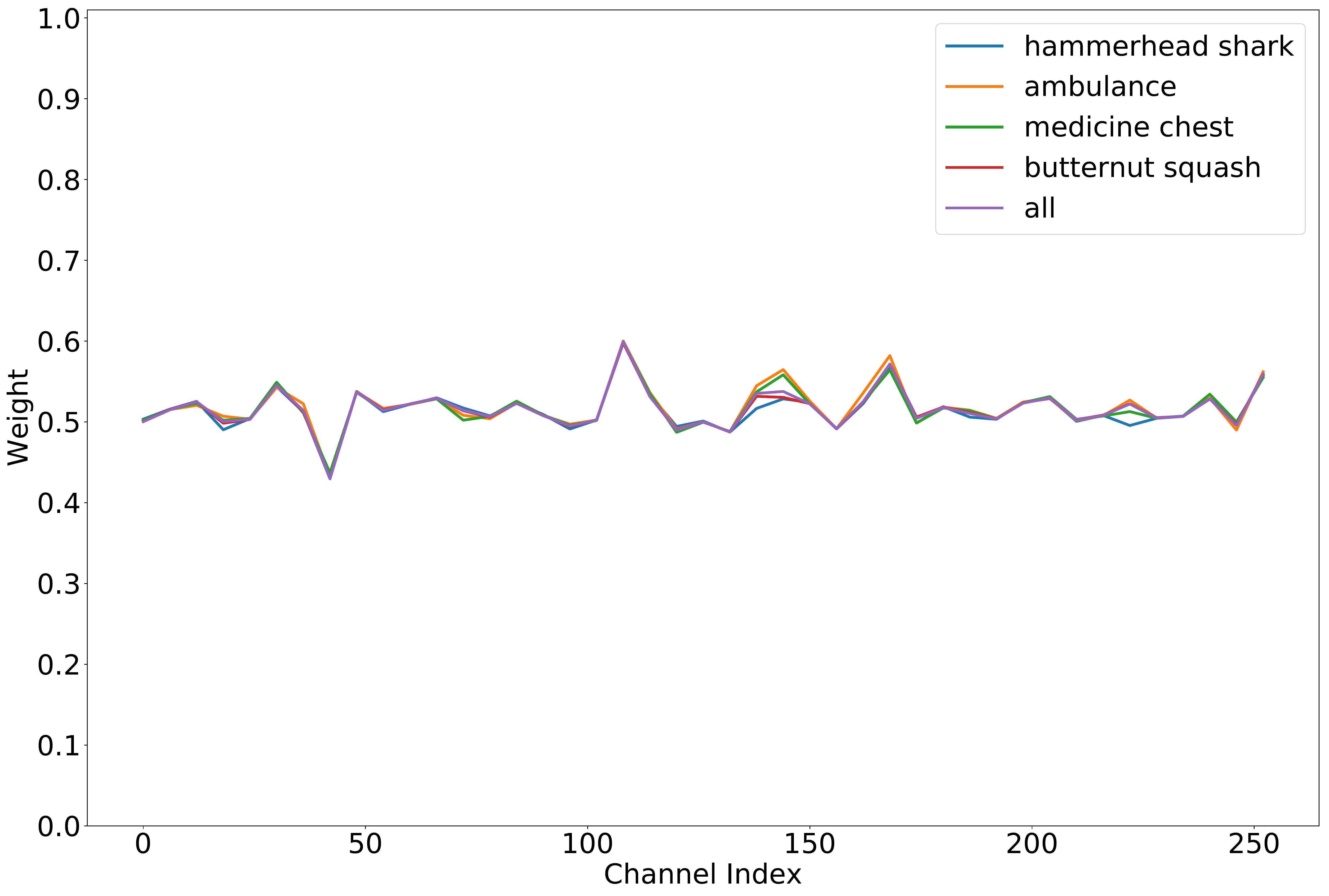}
		\end{minipage}
	}%
	\subfigure[conv\_3\_1]{
		\begin{minipage}[t]{0.25\linewidth}
			\centering
			\includegraphics[width=0.8\columnwidth]{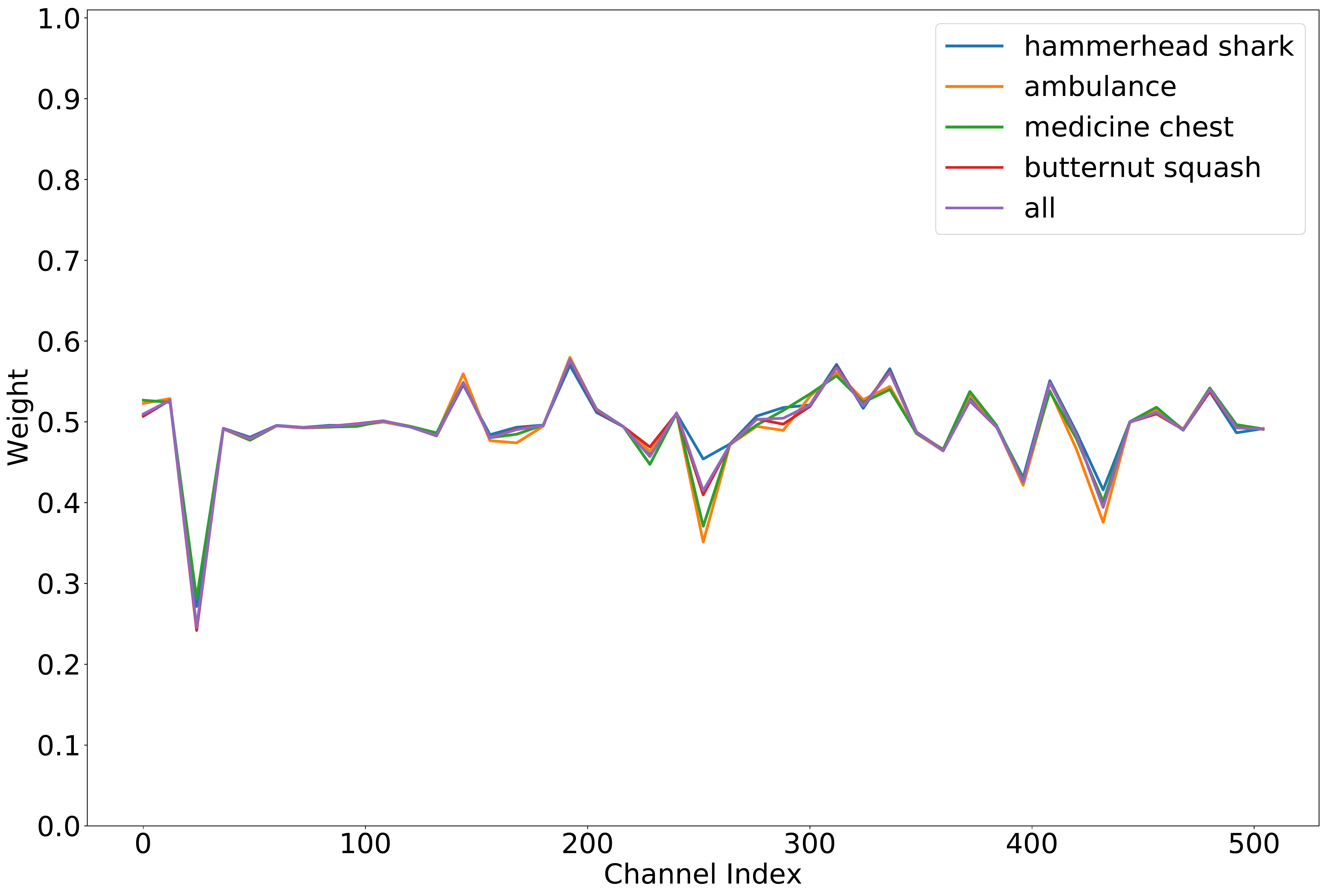}
			\includegraphics[width=0.8\columnwidth]{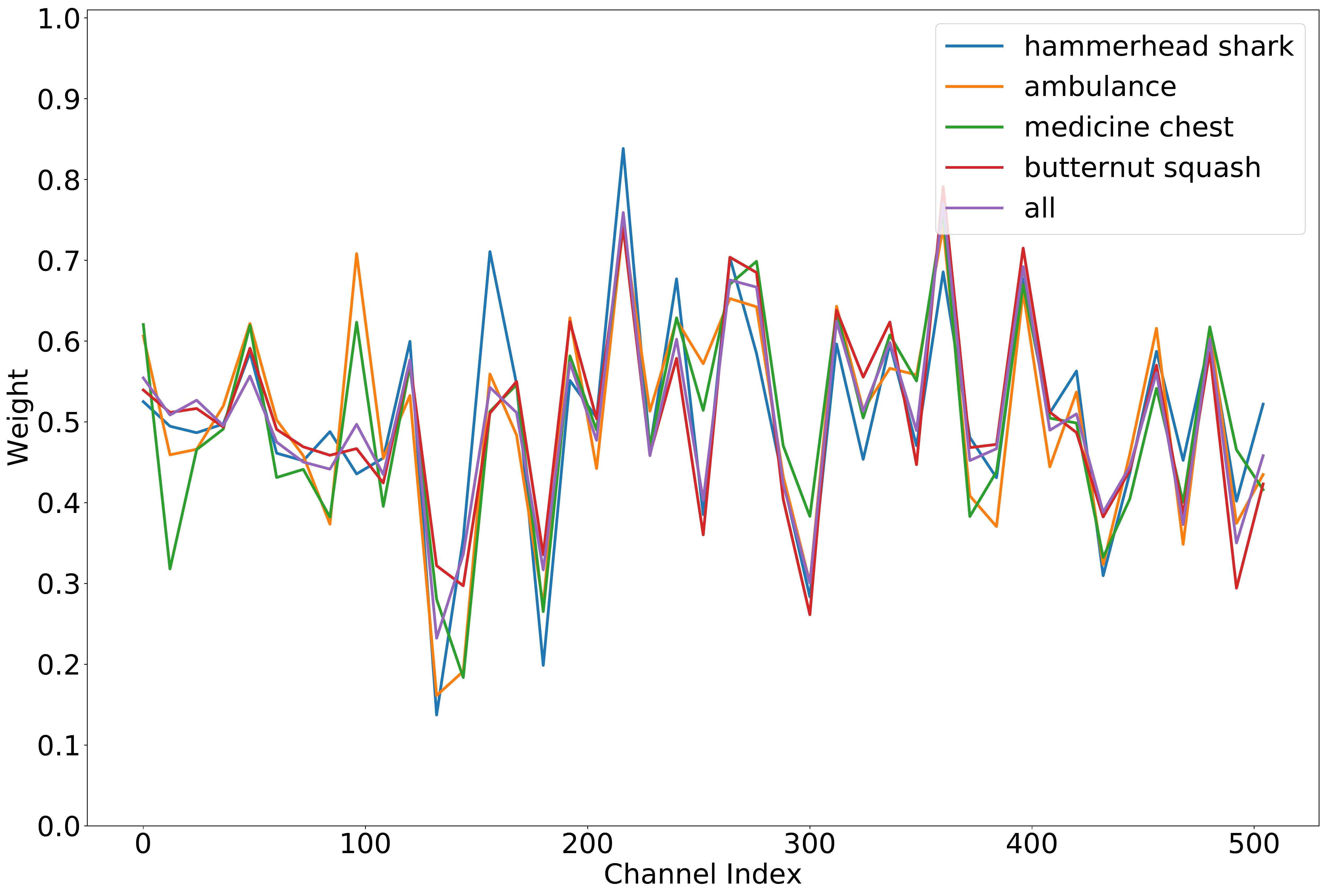}
		\end{minipage}
	}%
	
	\subfigure[conv\_3\_2]{
		\begin{minipage}[t]{0.25\linewidth}
			\centering
			\includegraphics[width=0.8\columnwidth]{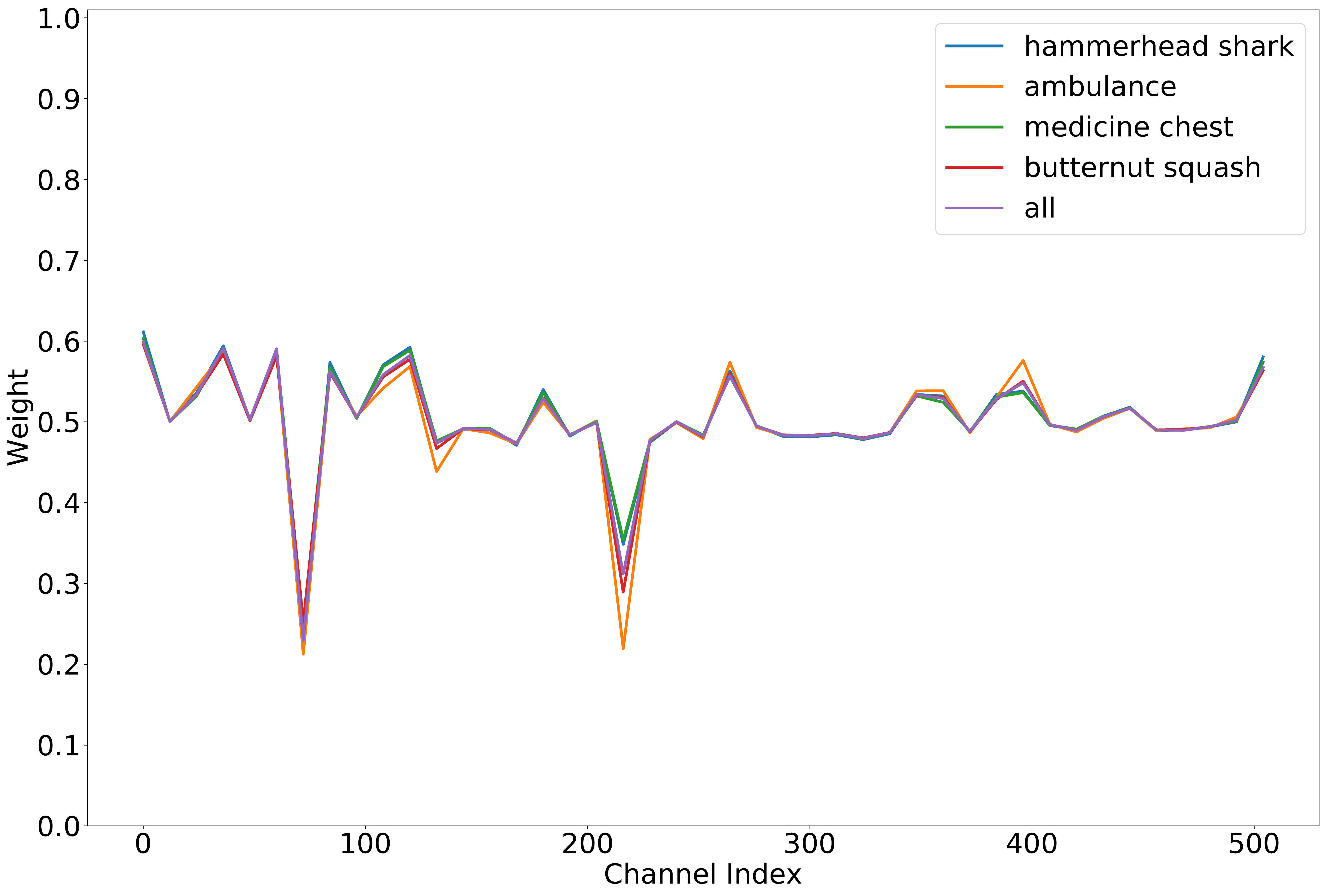}
			\includegraphics[width=0.8\columnwidth]{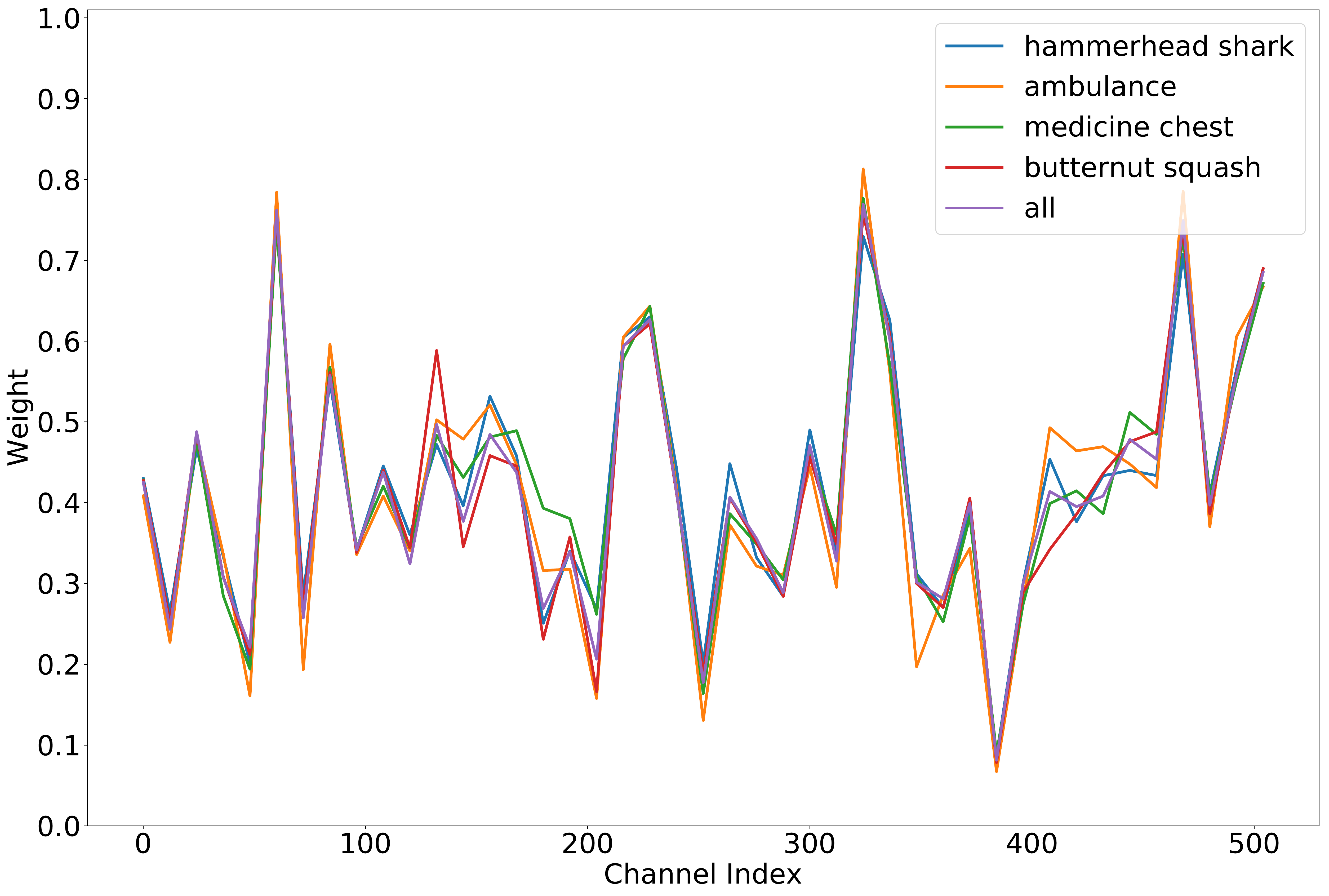}
		\end{minipage}%
	}%
	\subfigure[conv\_3\_3]{
		\begin{minipage}[t]{0.25\linewidth}
			\centering
			\includegraphics[width=0.8\columnwidth]{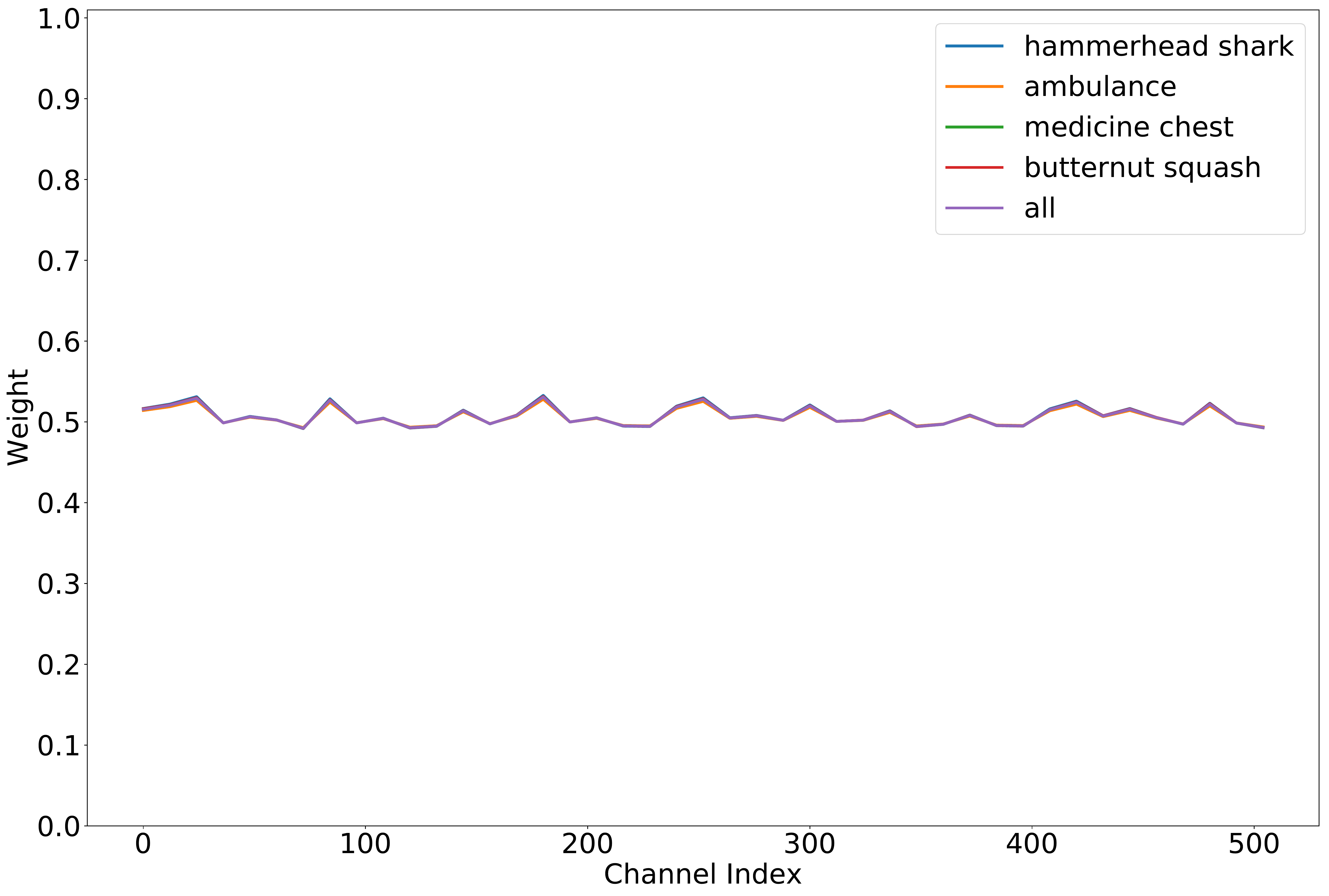}						\includegraphics[width=0.8\columnwidth]{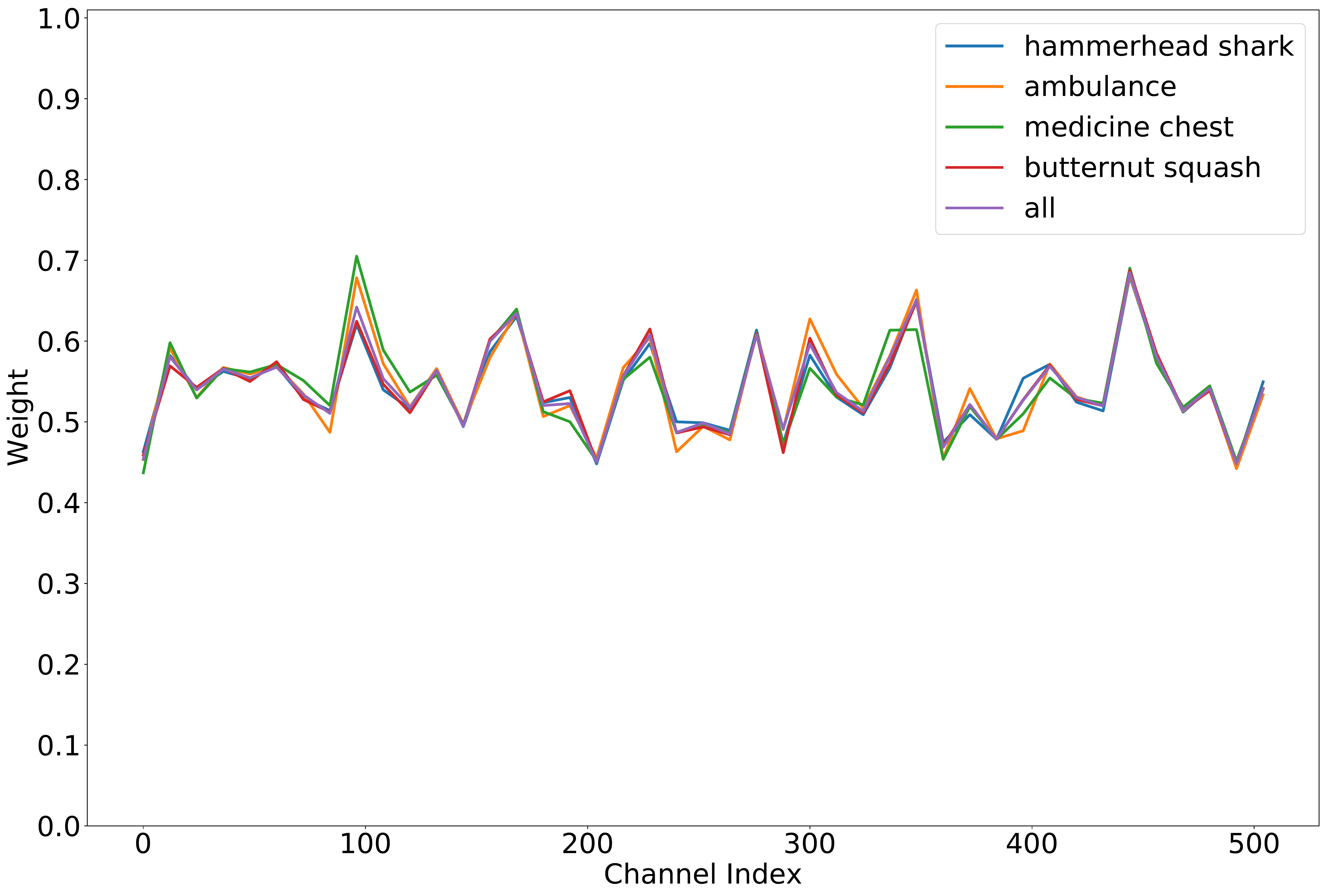}
		\end{minipage}%
	}%
	\subfigure[conv\_3\_4]{
		\begin{minipage}[t]{0.25\linewidth}
			\centering
			\includegraphics[width=0.8\columnwidth]{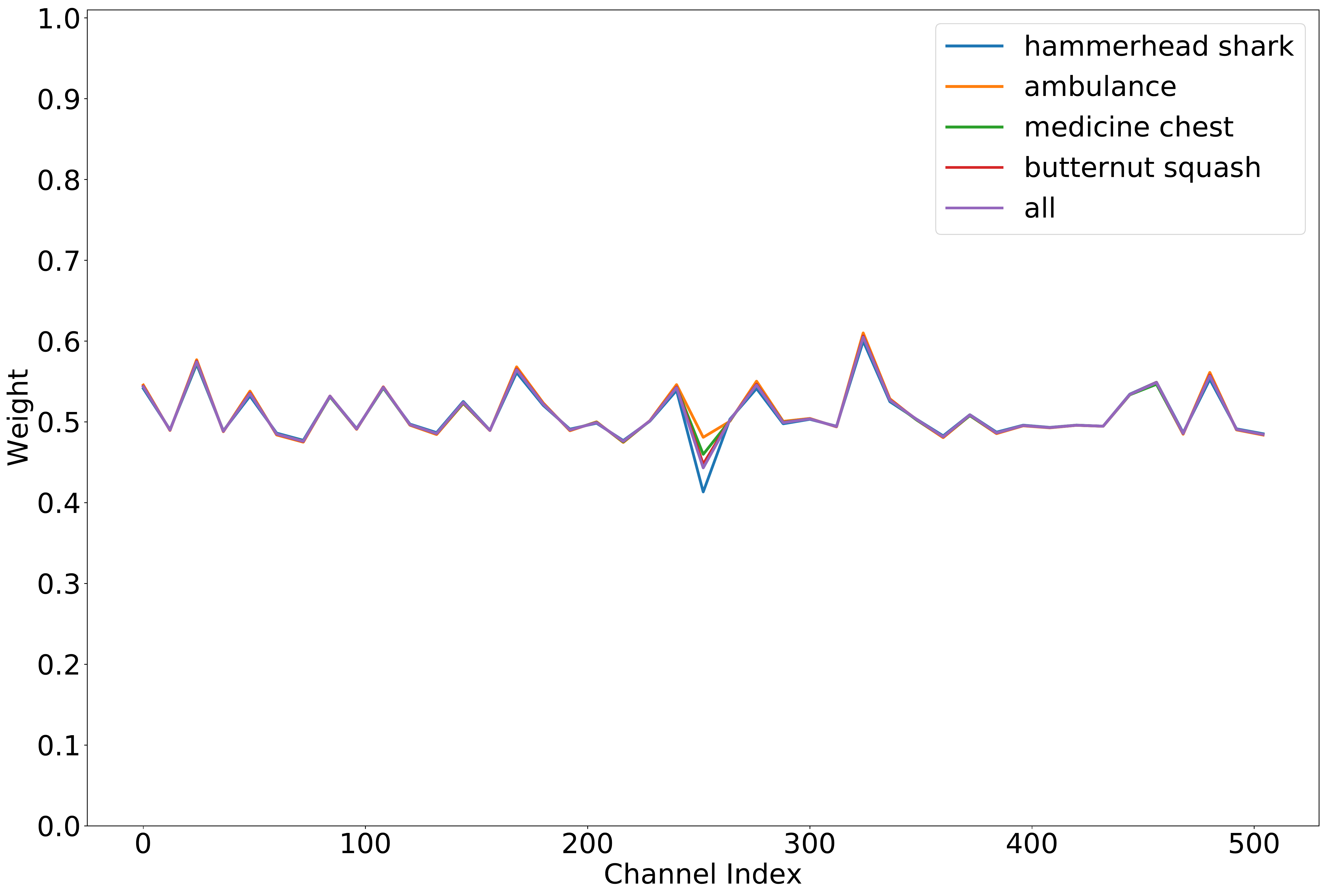}
			\includegraphics[width=0.8\columnwidth]{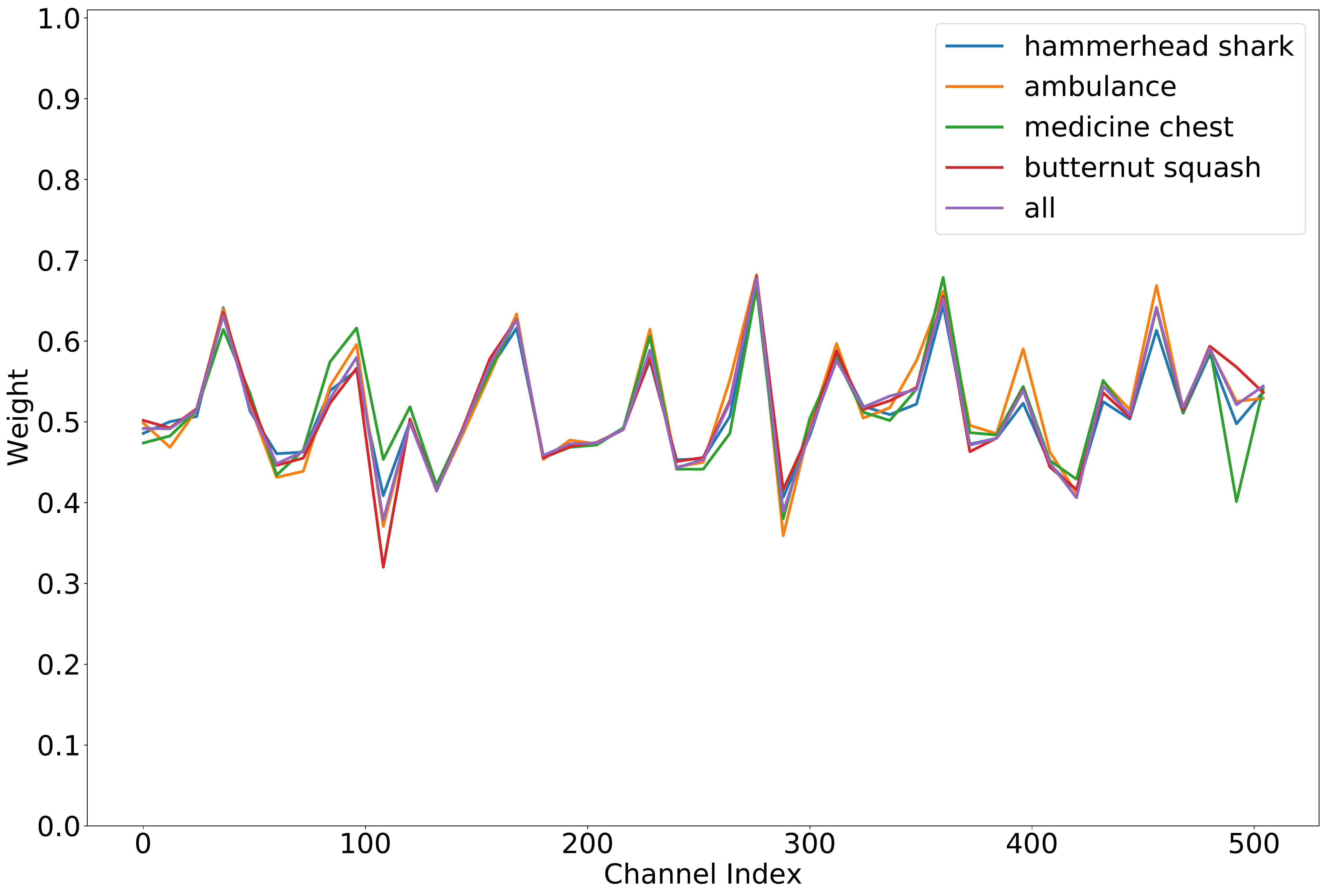}
		\end{minipage}
	}%
	\subfigure[conv\_4\_1]{
		\begin{minipage}[t]{0.25\linewidth}
			\centering
			\includegraphics[width=0.8\columnwidth]{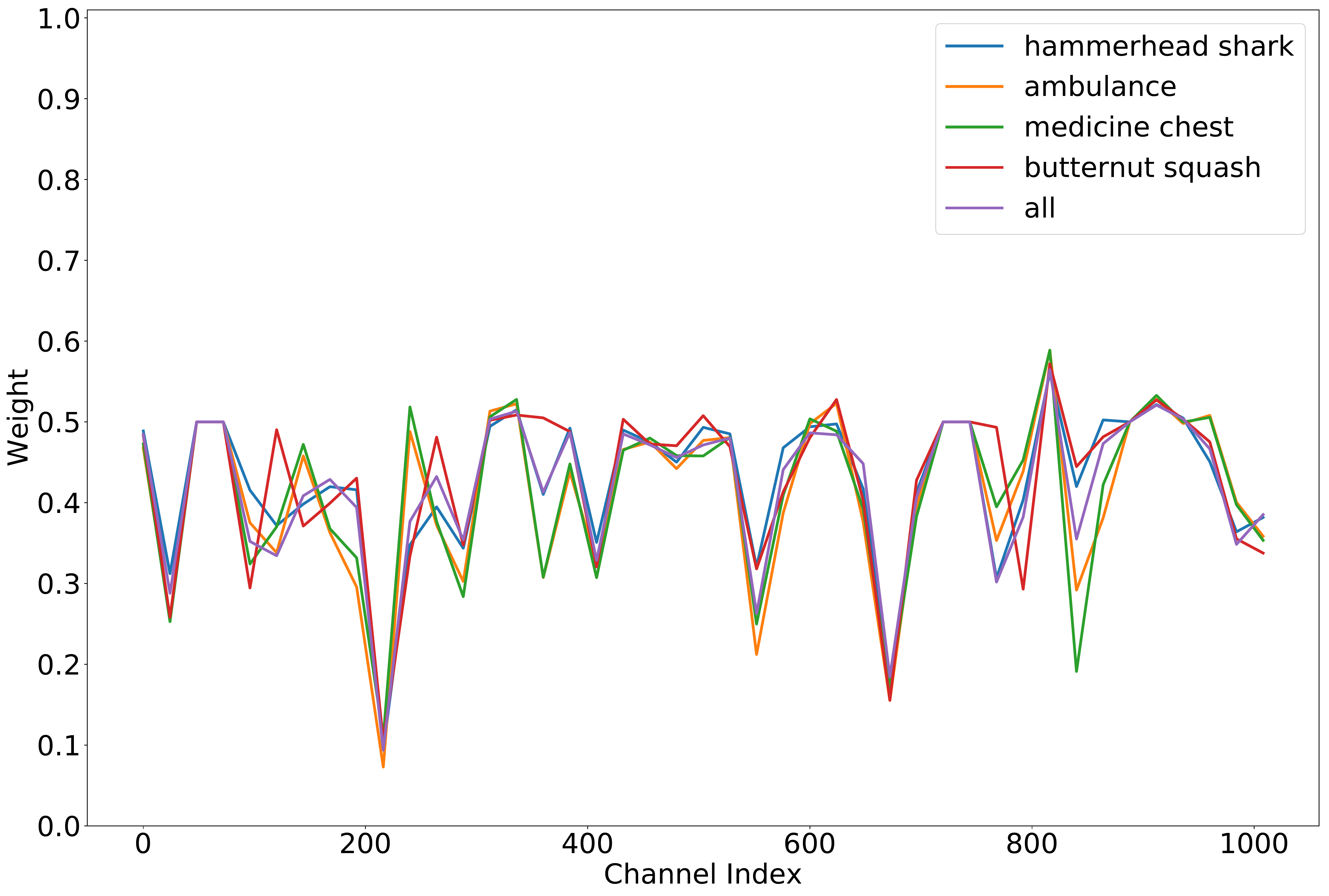}
			\includegraphics[width=0.8\columnwidth]{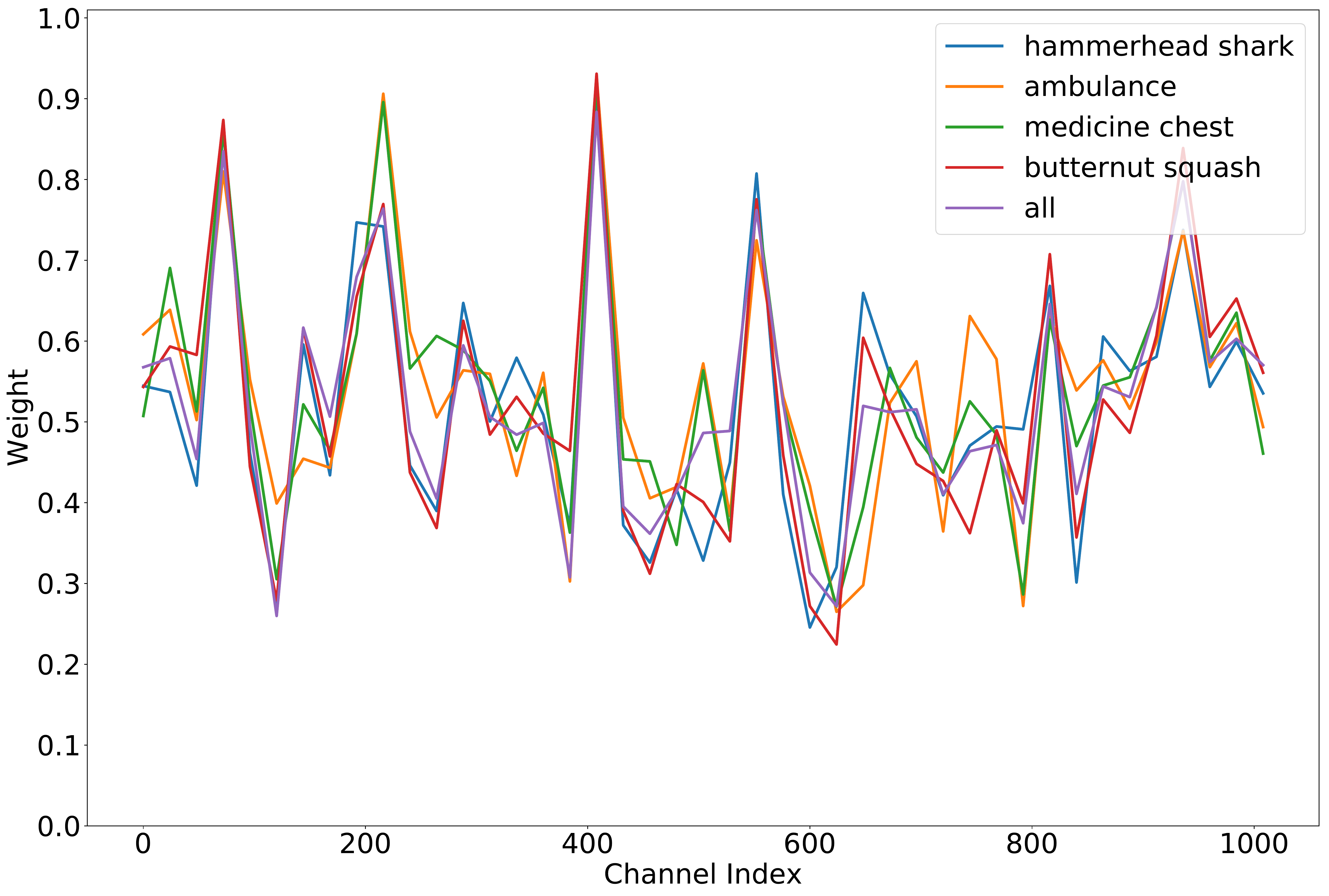}
		\end{minipage}
	}%
	
	\subfigure[conv\_4\_2]{
		\begin{minipage}[t]{0.25\linewidth}
			\centering
			\includegraphics[width=0.8\columnwidth]{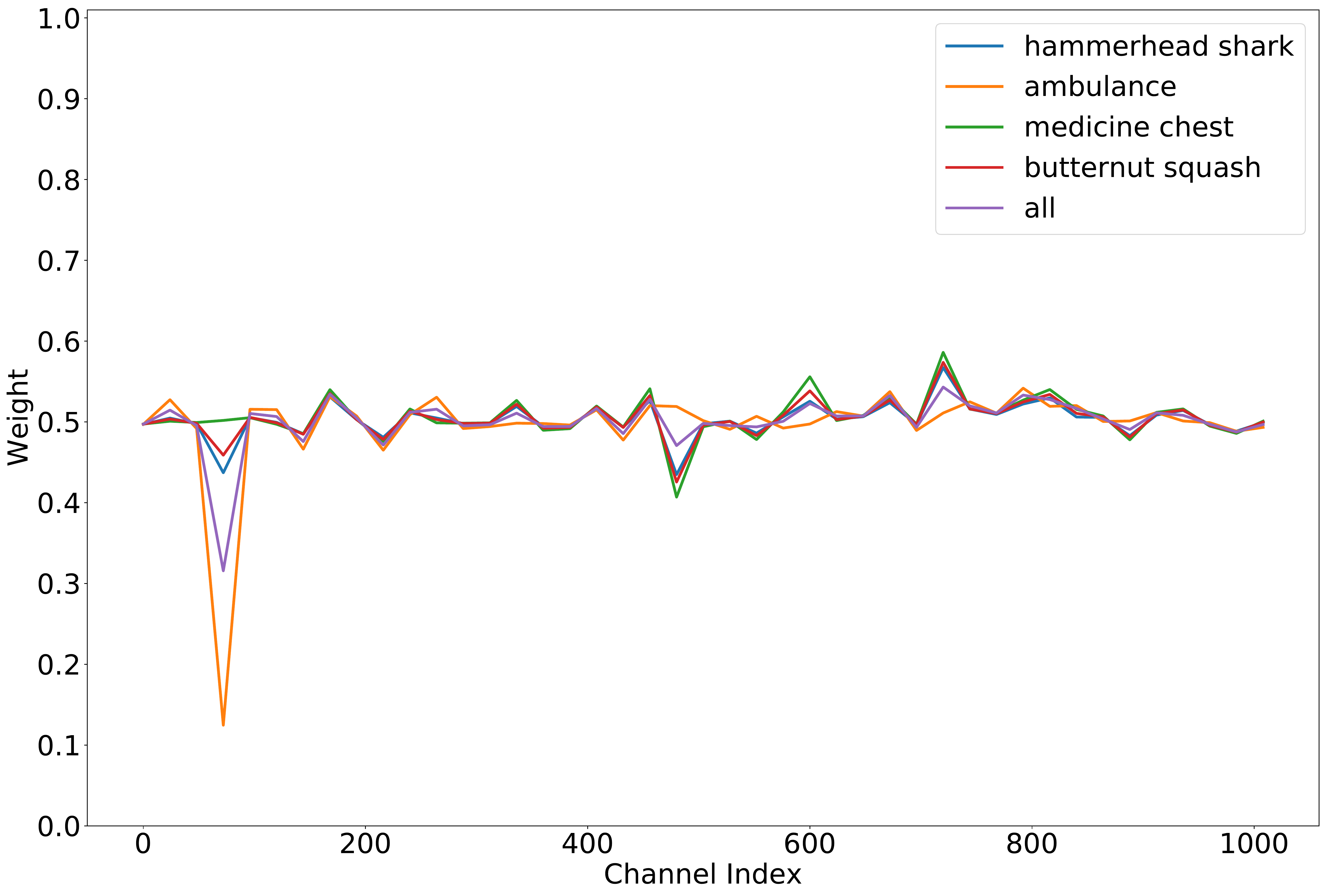}
			\includegraphics[width=0.8\columnwidth]{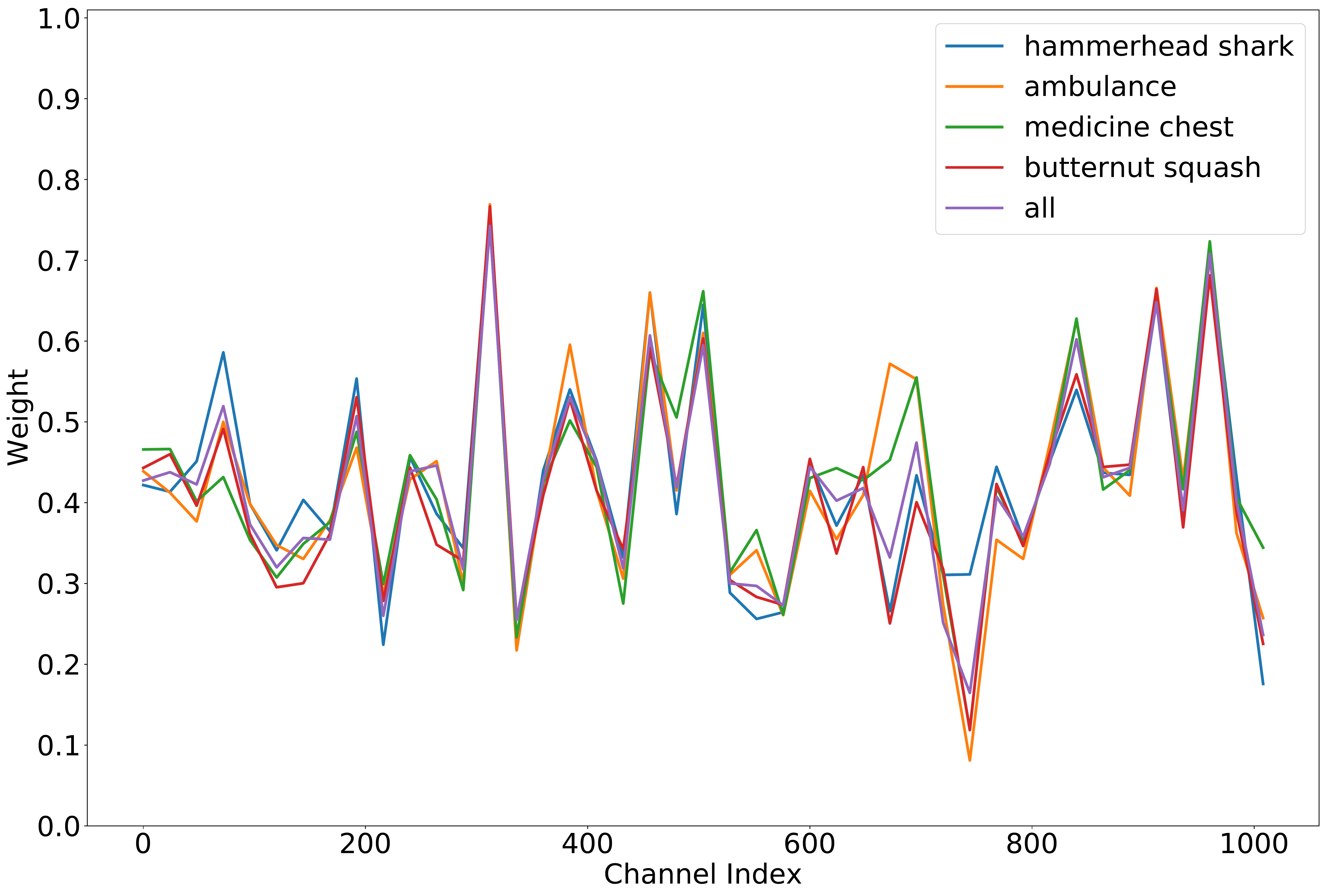}
		\end{minipage}%
	}%
	\subfigure[conv\_4\_3]{
		\begin{minipage}[t]{0.25\linewidth}
			\centering
			\includegraphics[width=0.8\columnwidth]{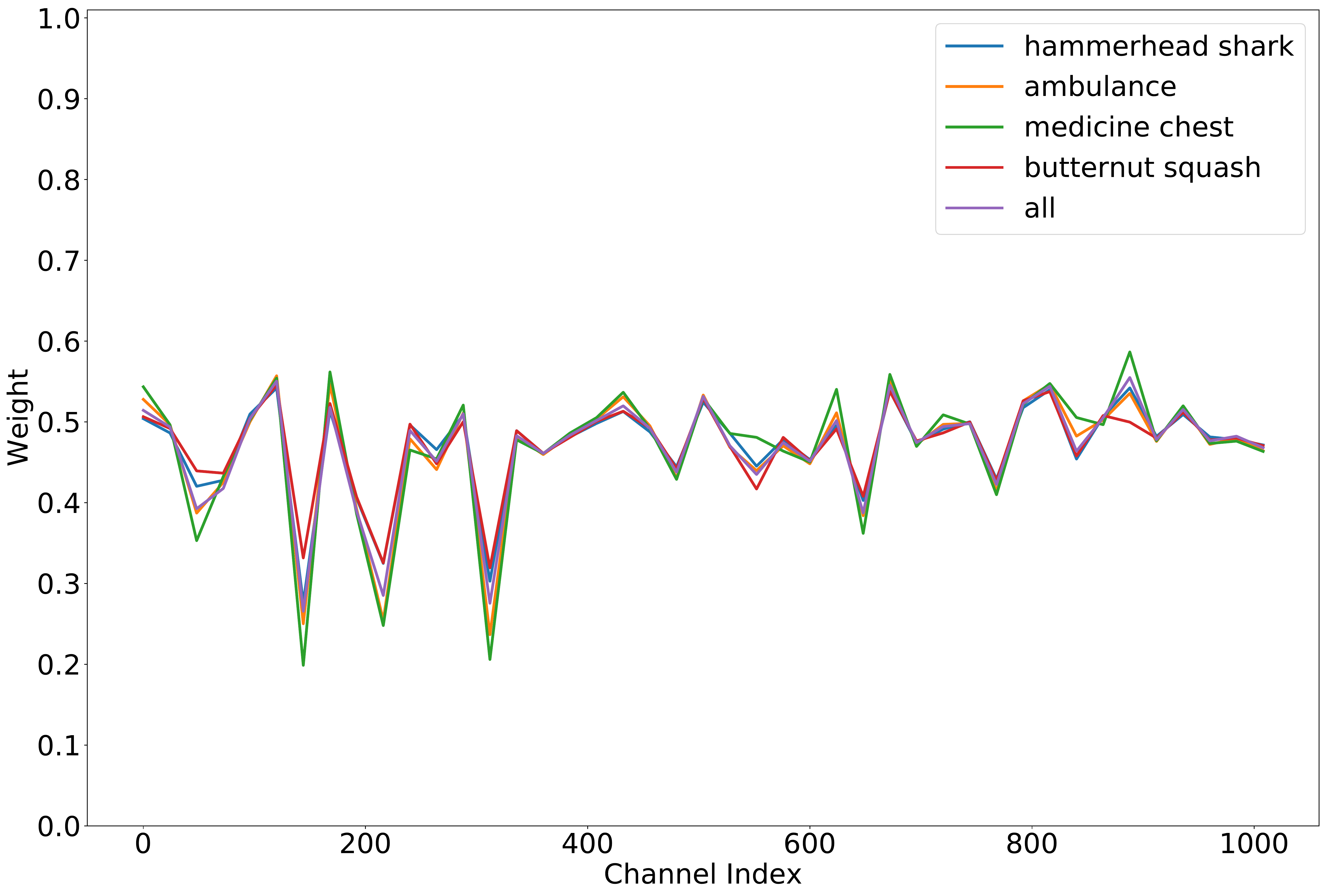}						\includegraphics[width=0.8\columnwidth]{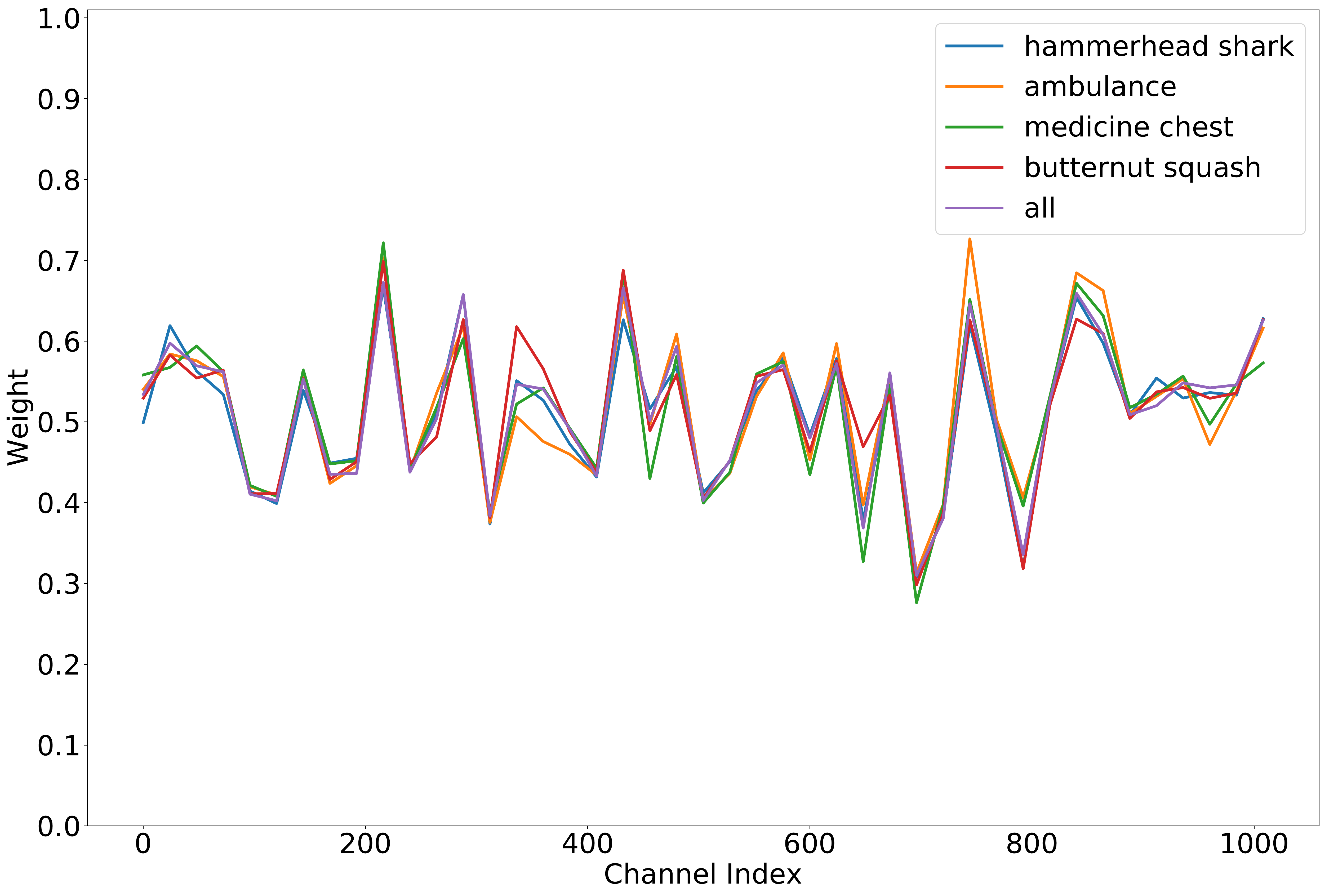}
		\end{minipage}%
	}%
	\subfigure[conv\_4\_4]{
		\begin{minipage}[t]{0.25\linewidth}
			\centering
			\includegraphics[width=0.8\columnwidth]{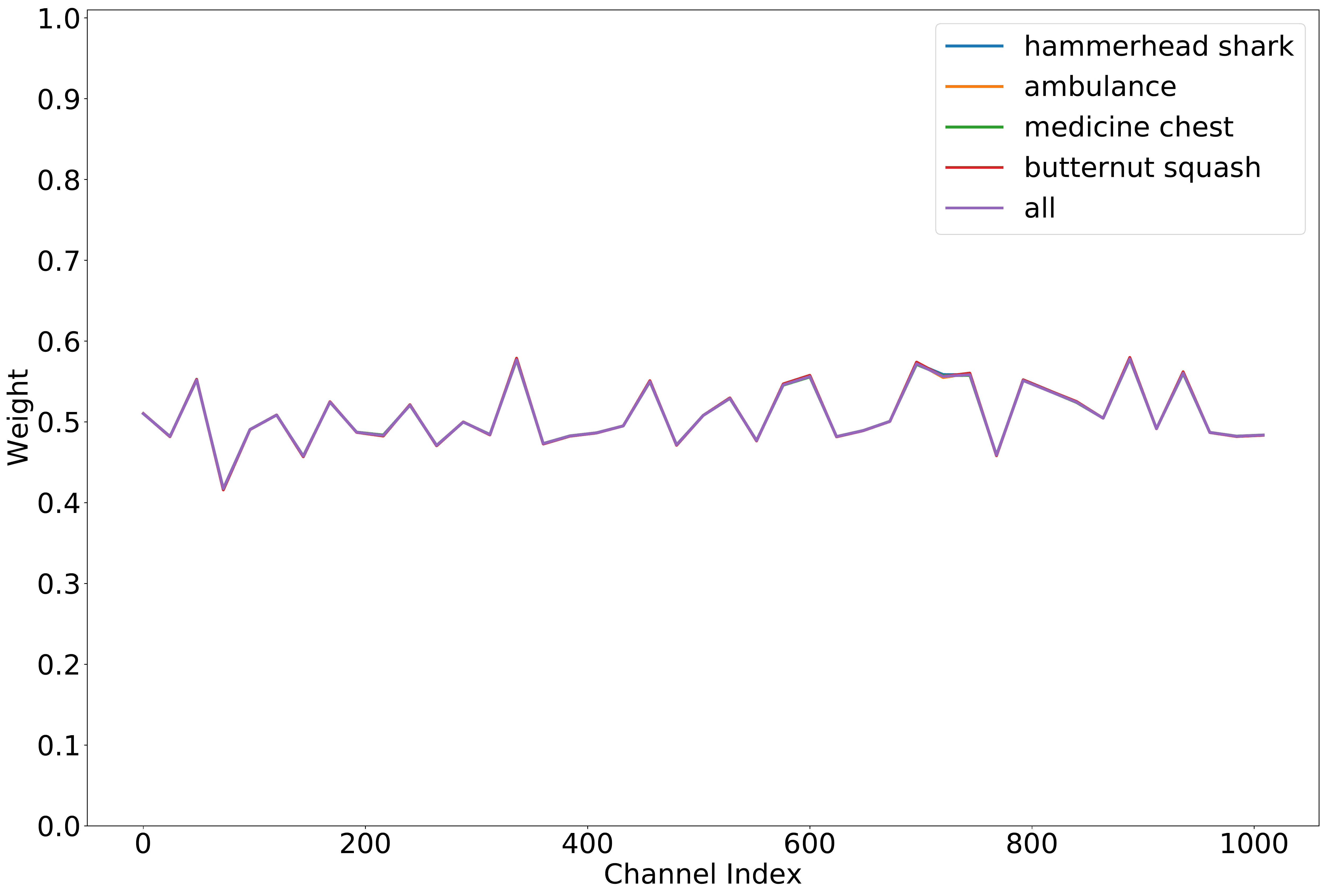}
			\includegraphics[width=0.8\columnwidth]{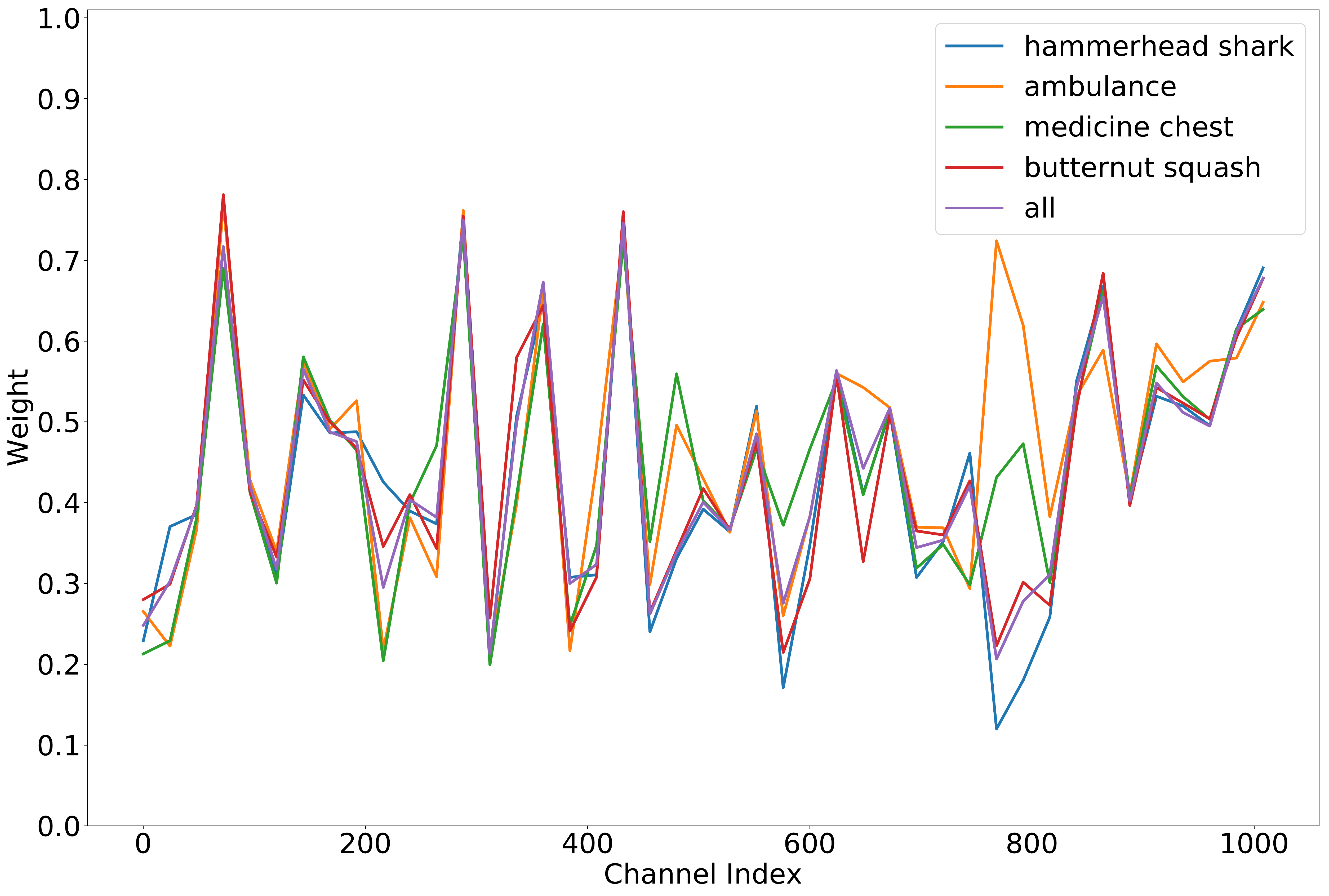}
		\end{minipage}
	}%
	\subfigure[conv\_4\_5]{
		\begin{minipage}[t]{0.25\linewidth}
			\centering
			\includegraphics[width=0.8\columnwidth]{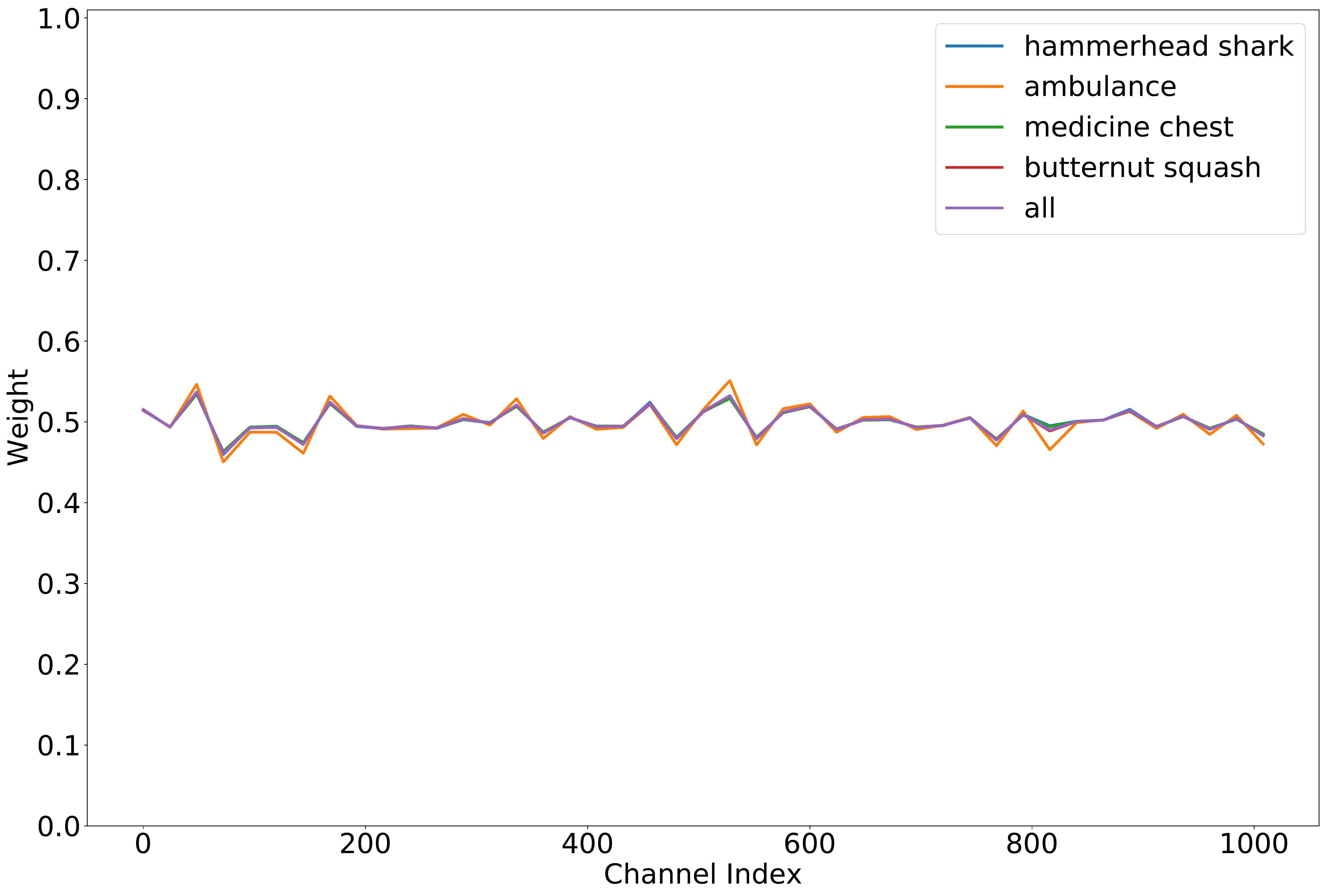}
			\includegraphics[width=0.8\columnwidth]{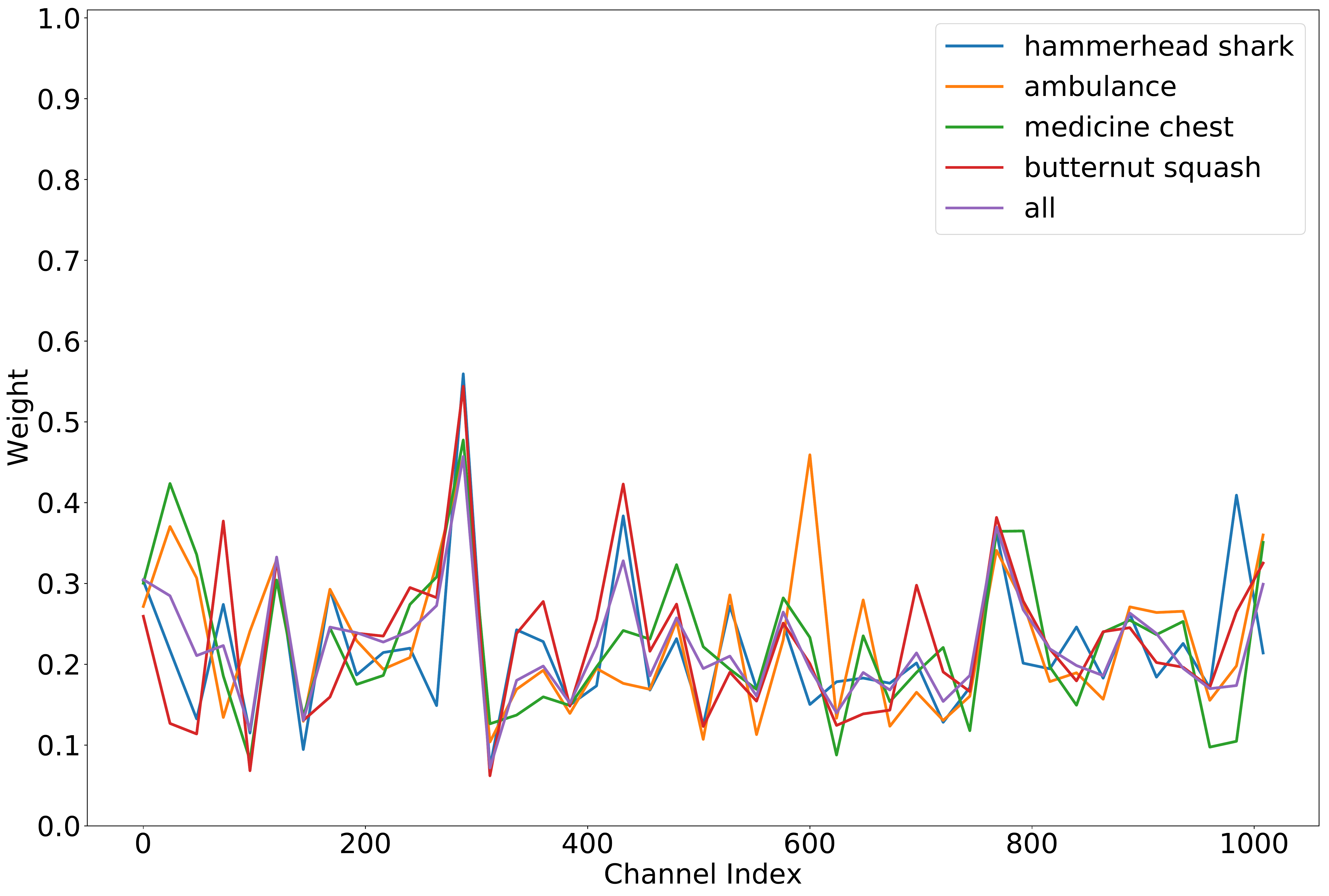}
		\end{minipage}
	}%
	
	\subfigure[conv\_4\_6]{
		\begin{minipage}[t]{0.25\linewidth}
			\centering
			\includegraphics[width=0.8\columnwidth]{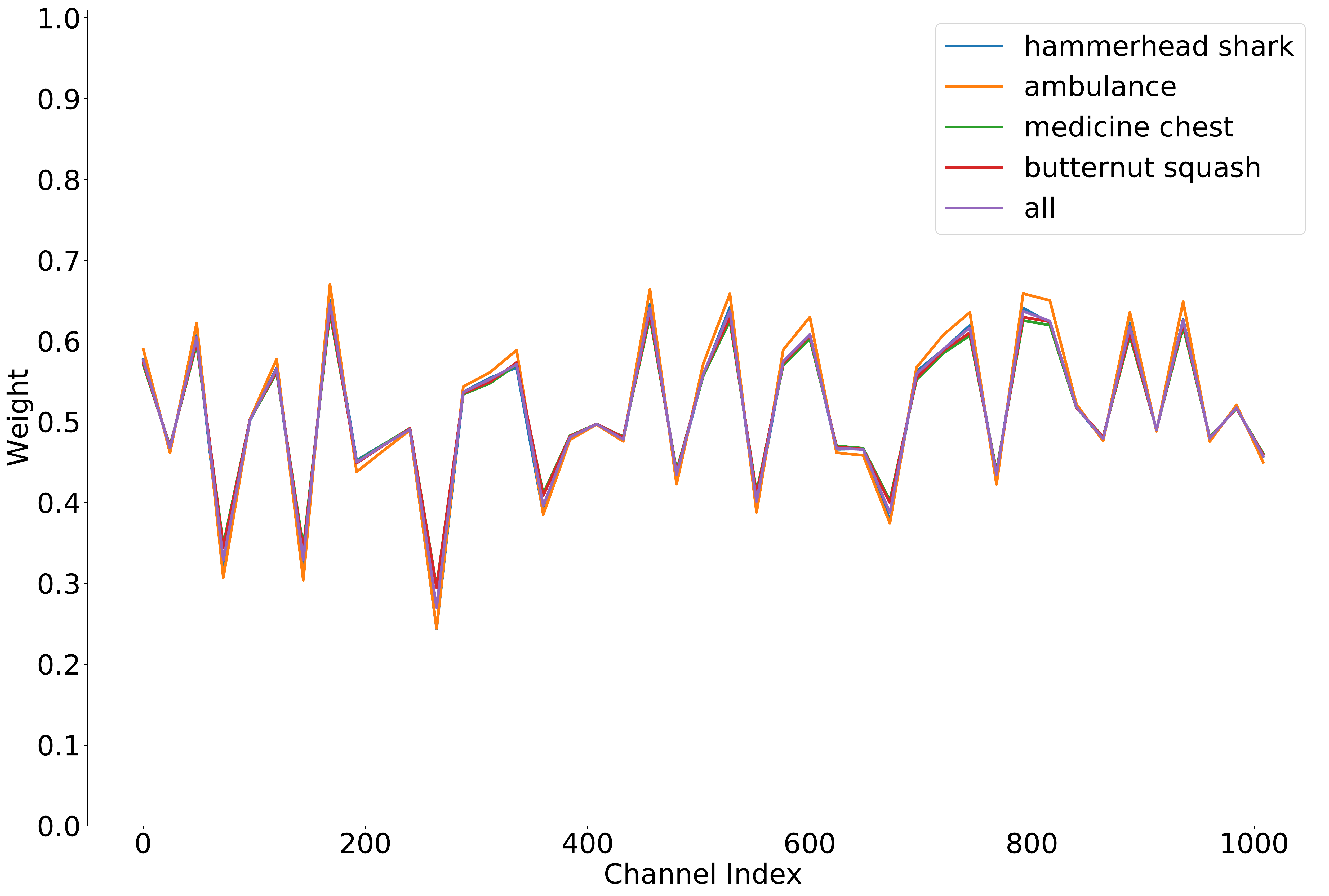}
			\includegraphics[width=0.8\columnwidth]{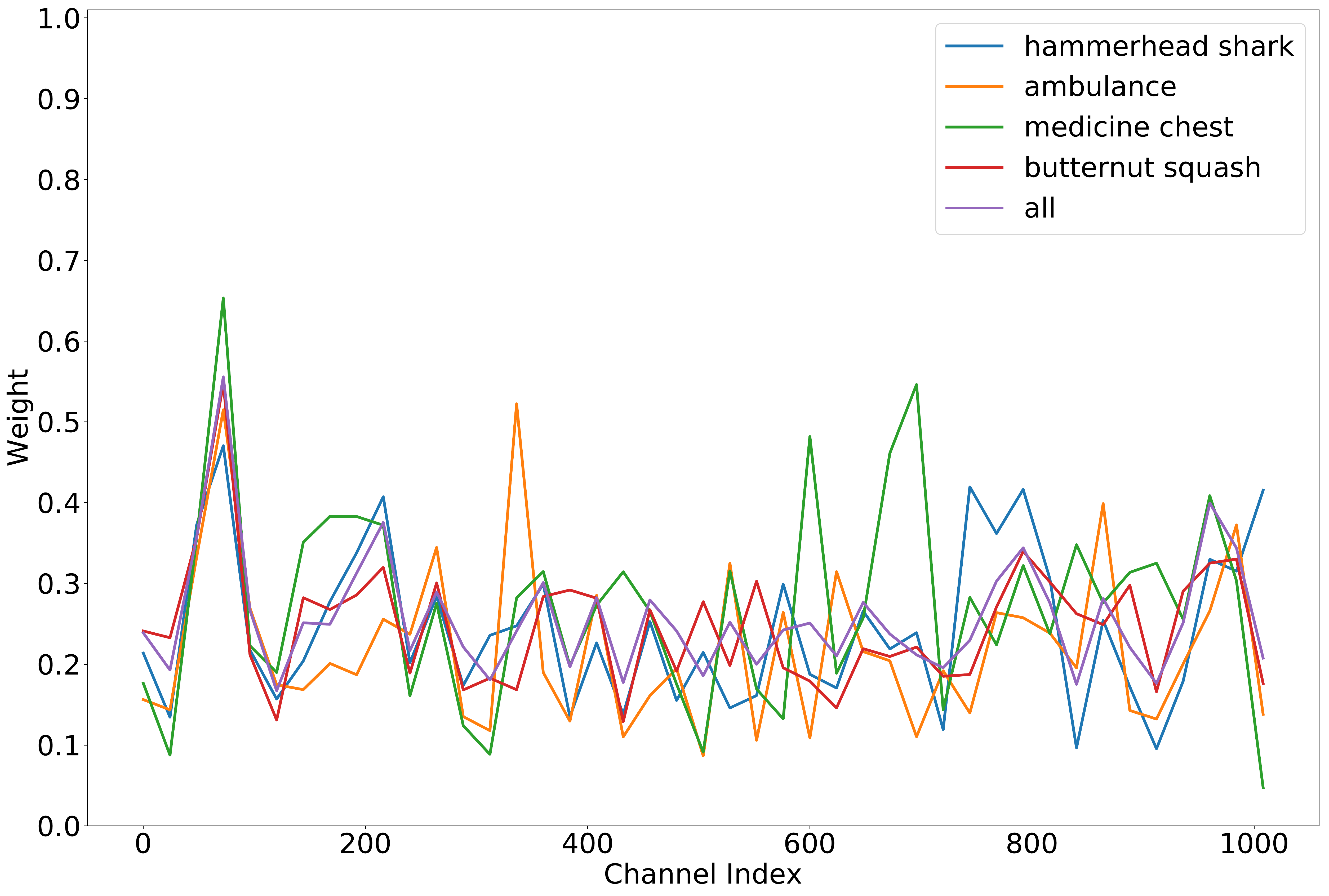}
		\end{minipage}%
	}%
	\subfigure[conv\_5\_1]{
		\begin{minipage}[t]{0.25\linewidth}
			\centering
			\includegraphics[width=0.8\columnwidth]{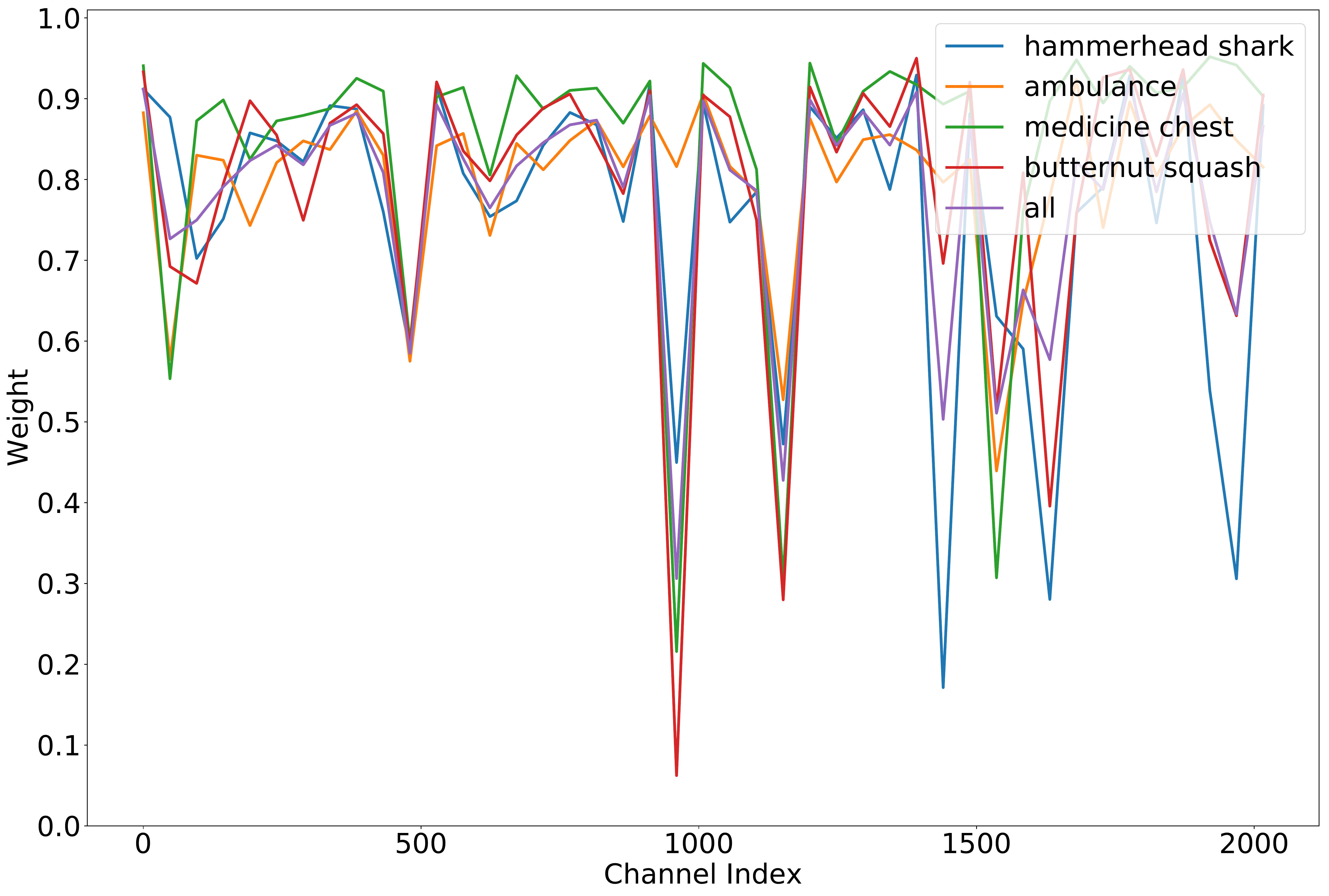}						\includegraphics[width=0.8\columnwidth]{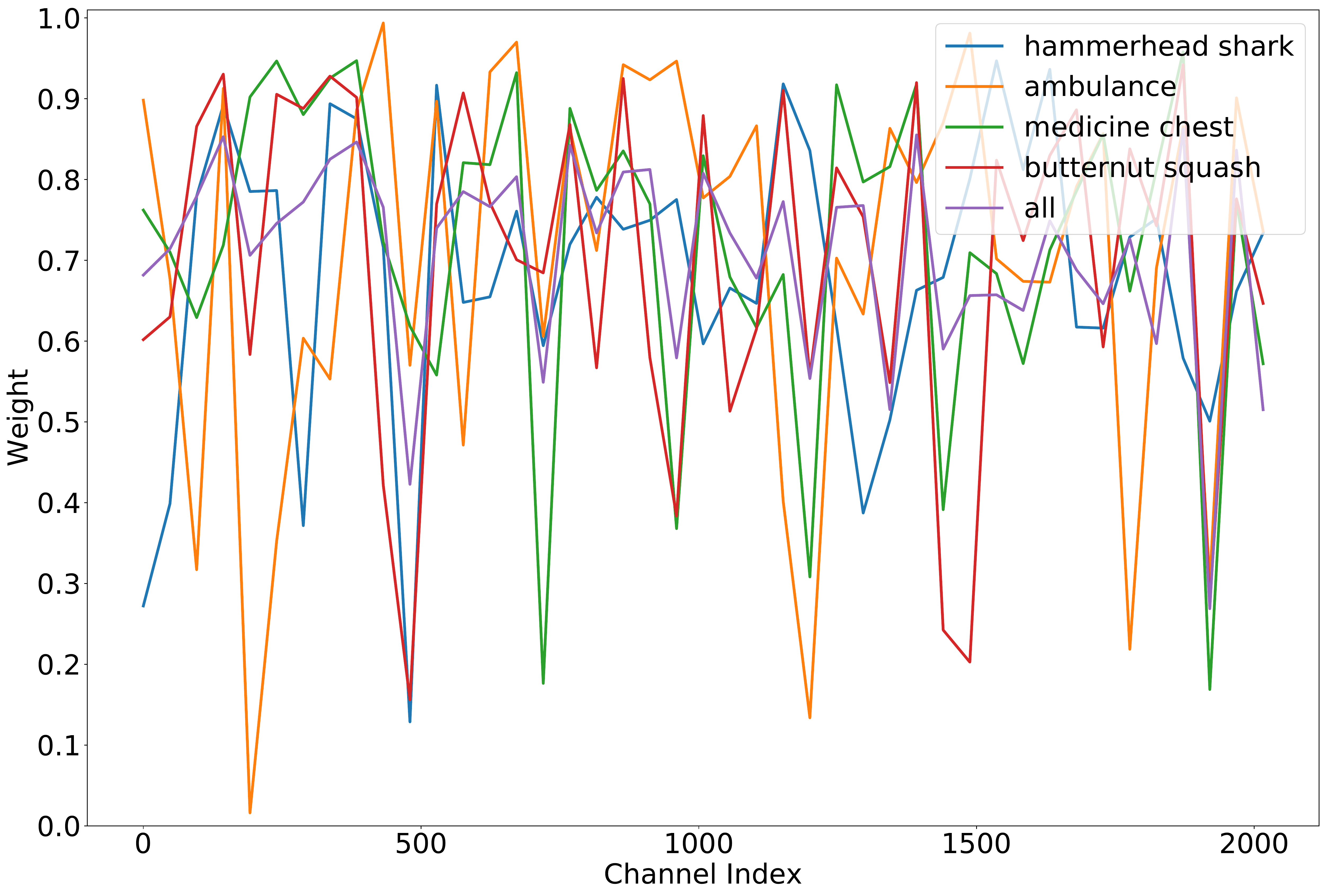}
		\end{minipage}%
	}%
	\subfigure[conv\_5\_2]{
		\begin{minipage}[t]{0.25\linewidth}
			\centering
			\includegraphics[width=0.8\columnwidth]{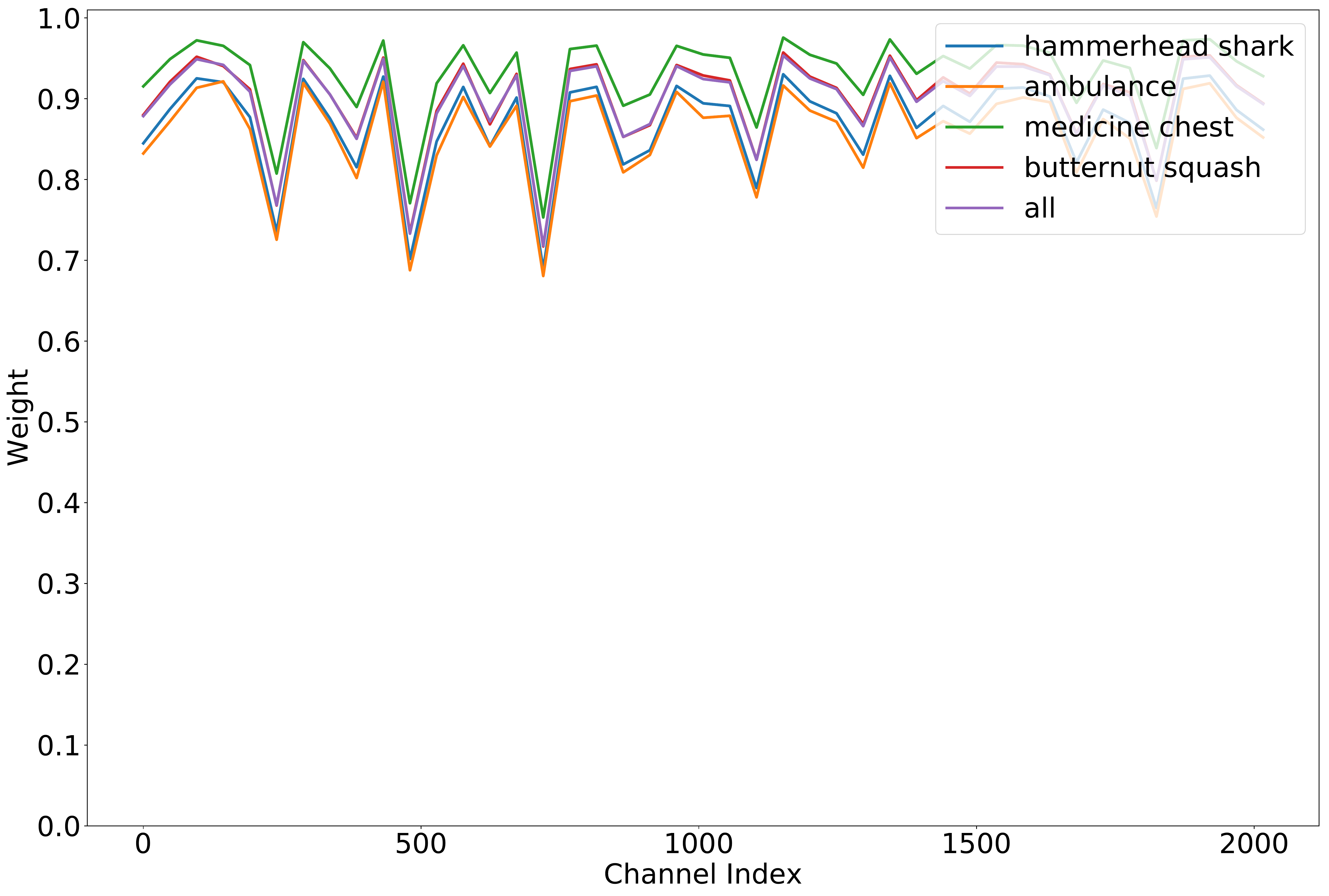}
			\includegraphics[width=0.8\columnwidth]{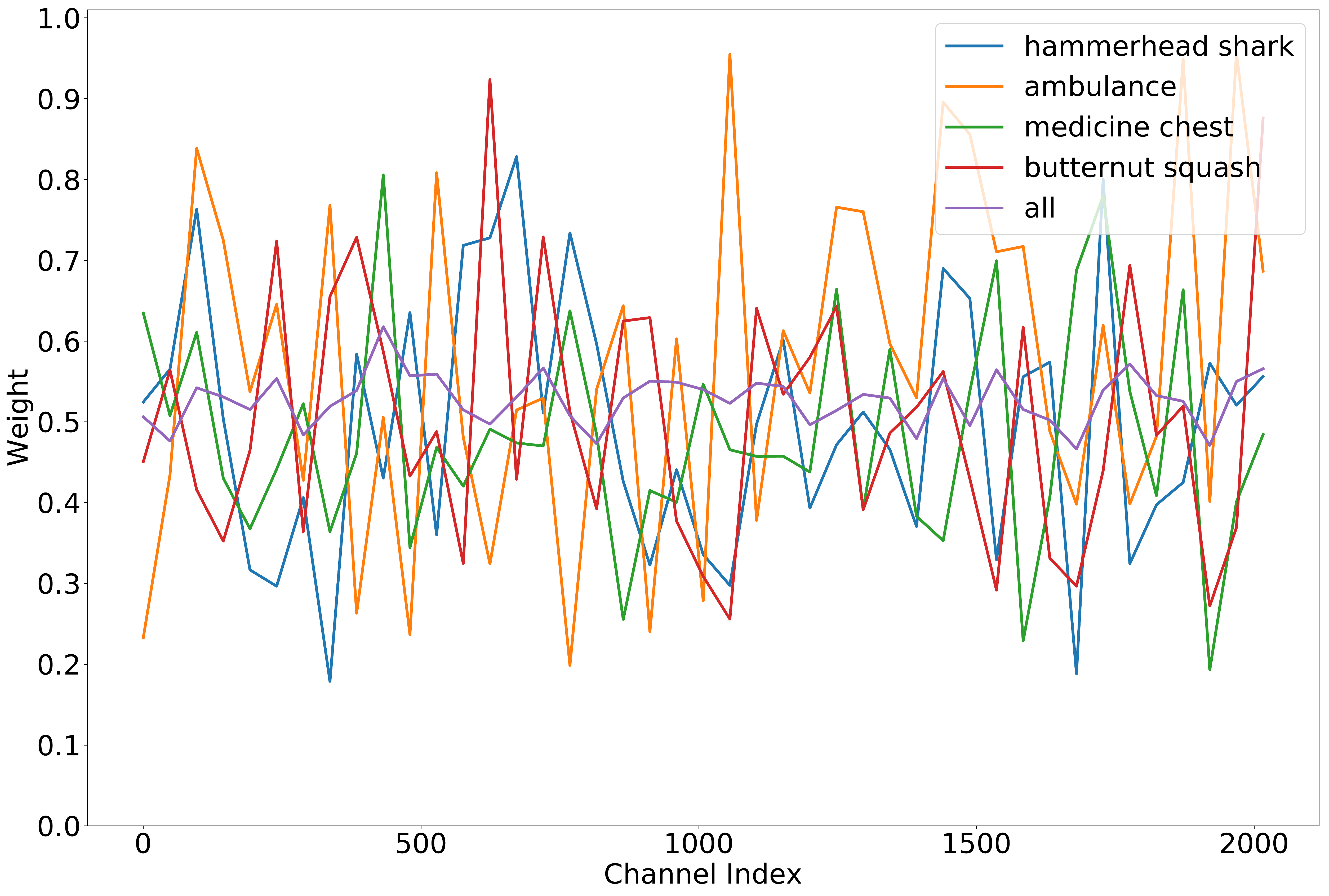}
		\end{minipage}
	}%
	\subfigure[conv\_5\_3]{
		\begin{minipage}[t]{0.25\linewidth}
			\centering
			\includegraphics[width=0.8\columnwidth]{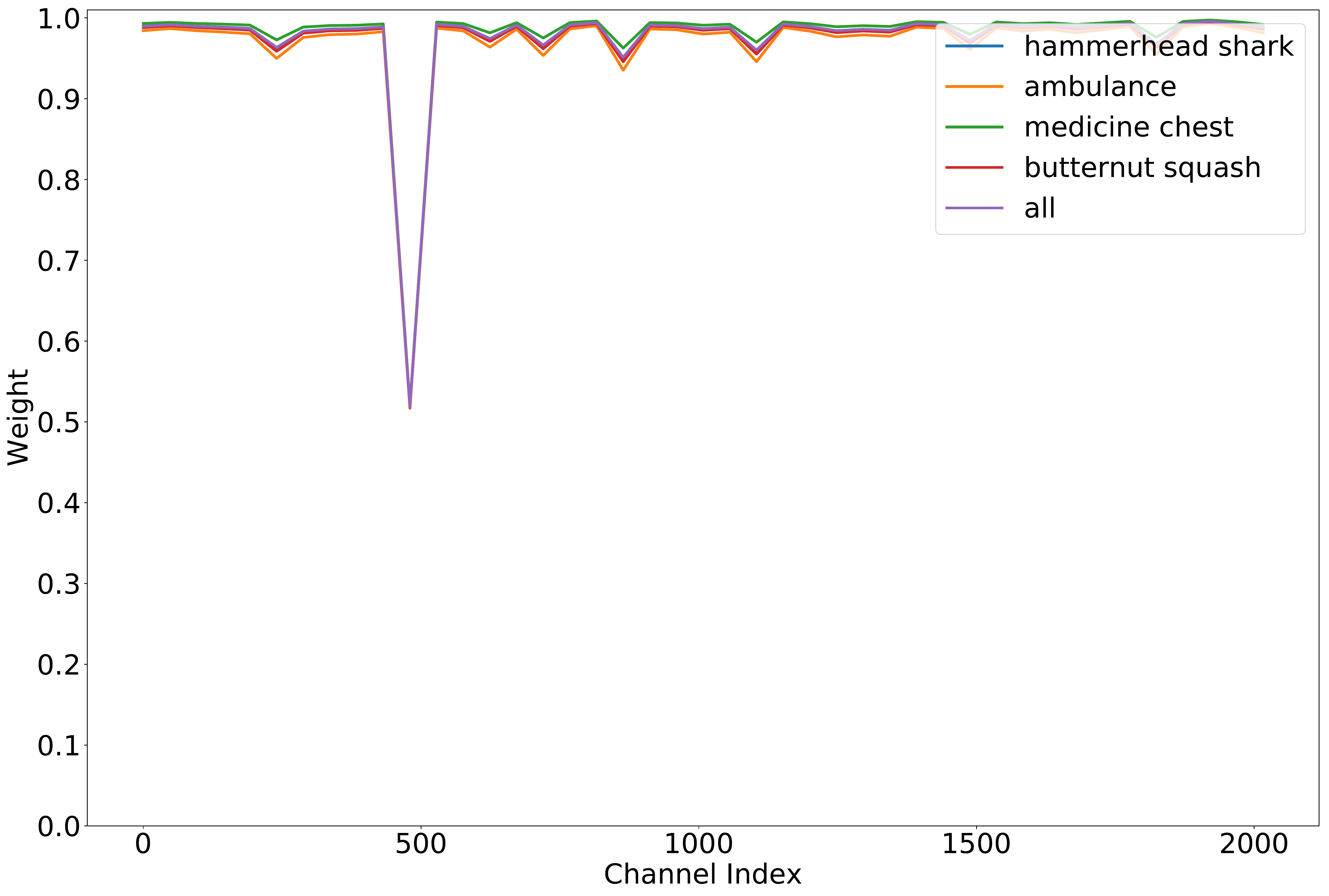}
			\includegraphics[width=0.8\columnwidth]{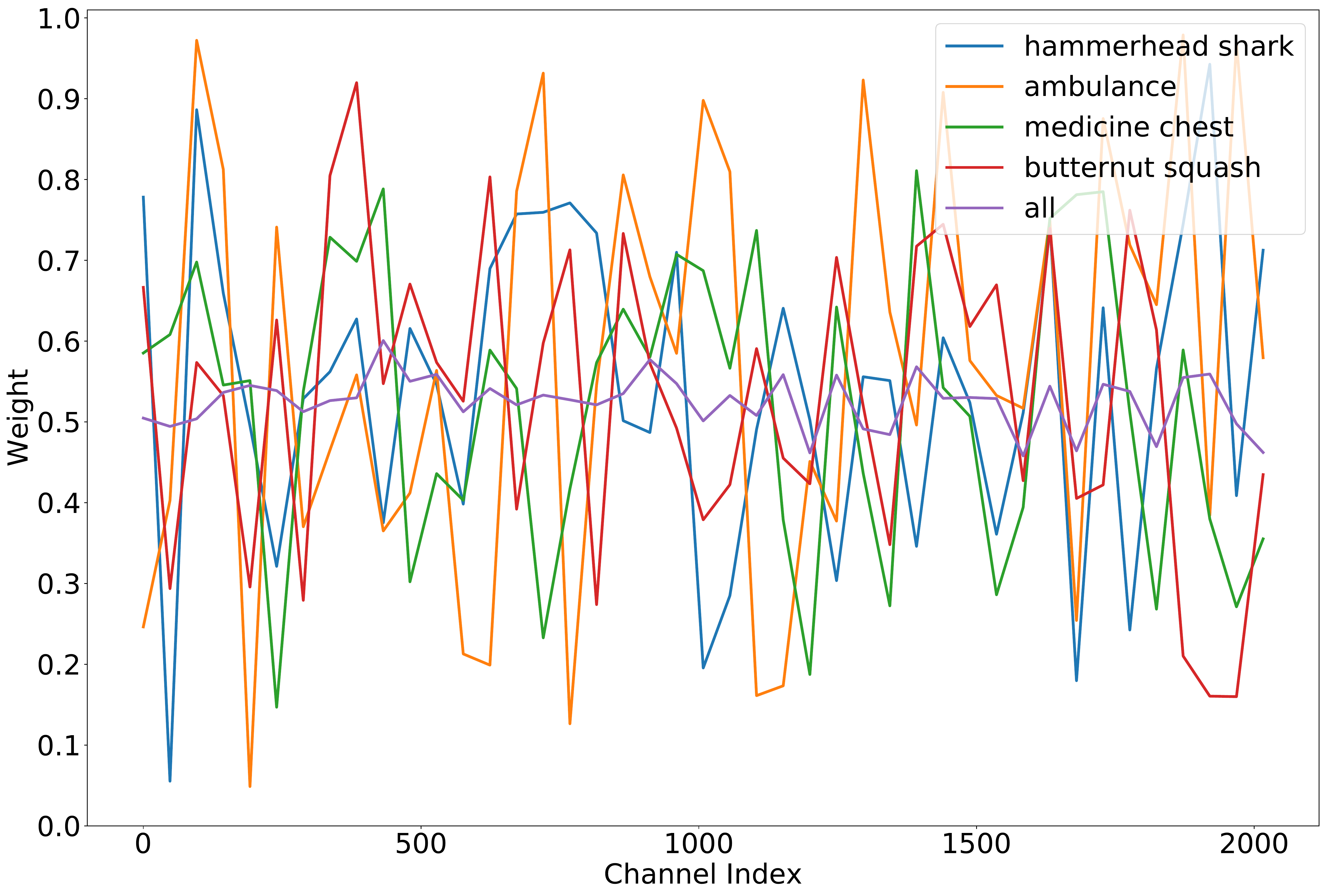}
		\end{minipage}
	}%
	\caption{Visualization the channel weights of conv\_$i$\_$j$, where $i$ indicate $i$-th stage and $j$ is $j$-th convolution block in $i$-th stage. The channel weights learned by ECA modules and SE blocks are illustrated in bottom and top of each row, respectively. Better view with zooming in.}
	\label{visualization}
\end{figure*}

\section*{Appendix III: Visualization of Weights Learned by ECA Modules and SE Blocks}
To further analyze the effect of our ECA module on learning channel attention, we visualize the weights learned by ECA modules and compare with SE blocks. Here, we employ ResNet-50 as backbone model, and illustrate weights of different convolution blocks. Specifically, we randomly sample four classes from ImageNet dataset, which are \textit{hammerhead shark}, \textit{ambulance}, \textit{medicine chest} and \textit{butternut squash}, respectively. Some example images are illustrated in Figure~\ref{examples}. After training the networks, for all images of each class collected from validation set of ImageNet, we compute the  channel weights of convolution blocks on average. Figure~\ref{visualization} visualizes the channel weights of conv\_$i$\_$j$, which  indicates $j$-th convolution block in $i$-th stage. Besides the visualization results of four random sampled classes, we also give the distribution of the average weights across $1K$ classes as reference. The channel weights learned by ECA modules and SE blocks are illustrated in bottom and top of each row, respectively.

From Figure~\ref{visualization} we have the following observations. Firstly, for both ECA modules and SE blocks, the distributions of channel weights for different classes are very similar at the earlier layers (i.e., ones from conv\_2\_1 to conv\_3\_4), which may be by reason of that the earlier layers aim at capturing the basic elements (e.g., boundaries and corners)~\cite{DBLP:conf/eccv/ZeilerF14}. These features are almost similar for different classes. Such phenomenon also was described in the extended version of~\cite{SENet18}\footnote{https://arxiv.org/abs/1709.01507}. Secondly, for the channel weights of different classes learned by SE blocks, most of them tend to be the same (i.e., 0.5) in conv\_4\_2 $\sim$ conv\_4\_5  while the differences among various classes are not obvious. On the contrary, the weights learned by ECA modules are clearly different across various channels and classes. Since convolution blocks in 4-$th$ stage prefer to learn semantic information, so the weights learned by ECA modules can better distinguish different classes. Finally, convolution blocks in the final stage (i.e., conv\_5\_1, conv\_5\_2 and conv\_5\_3) capture high-level semantic features and they are more class-specific. Obviously, the weights learned by ECA modules are more class-specific than ones learned by SE blocks. Above results clearly demonstrate that the weights learned by our ECA modules have better discriminative ability.

\end{document}